 \documentclass[final,1p,times,authoryear]{elsarticle_mod}

\usepackage{amssymb}

\usepackage[utf8]{inputenc}

\journal{}

\usepackage[dvipsnames]{xcolor}
\usepackage{subcaption}
\usepackage{mathtools}
\usepackage{tikz}
\usepackage{pgf}
\usepackage{pgfplots}
\usepgfplotslibrary{groupplots}
\usetikzlibrary{bayesnet}
\usepackage[bottom]{footmisc}

\usepackage{placeins}

\definecolor{color0}{HTML}{485899}
\definecolor{color1}{HTML}{c25955}
\definecolor{color2}{HTML}{6c736c}

\pgfplotsset{
table/search path={gfx/data},
}

\newcommand\inputpgf[2]{{
\let\pgfimageWithoutPath\pgfimage
\renewcommand{\pgfimage}[2][]{\pgfimageWithoutPath[##1]{#1/##2}}
\input{#1/#2}
}}

\newcommand{\setalgorithmspacing}{\setstretch{1.25}}

\newcommand{\LS}{LS}

\newcommand{\resamplingstatement}{Results have been averaged by repeatedly training the model on resampled data.}

\newlength\figHeight
\newlength\figWidth
\DeclareUnicodeCharacter{2212}{-}

\newcommand{\flux}{\bm{J}}

\newcommand{\thetamap}{\bm{\theta}_{MAP}}

\newcommand{\vect}[1]{\bm{#1}}
\newcommand{\mat}[1]{\bm{#1}}

\newcommand{\diff}{\mathop{}\!\mathrm{d}}

\newcommand{\labeled}{l}
\newcommand{\unlabeled}{u}
\newcommand{\vo}{\mathcal{O}}
\newcommand{\obs}{\vect{\hat{o}}}

\usepackage{amssymb}
\usepackage{latexsym}
\usepackage{amsmath}
\usepackage{bbold}      
\usepackage[algoruled, linesnumbered]{algorithm2e}
\usepackage{diagbox}

\usepackage{url}
\usepackage{hyperref}
\usepackage{bm}
\usepackage{xcolor}
\usepackage{algorithm2e}
\usepackage{setspace} 
\usepackage{import}
\hypersetup{
    colorlinks,
    linkcolor={red!50!black},
    citecolor={blue!50!black},
    urlcolor={blue!80!black}
}
\usepackage{enumerate}
\usepackage{sidecap}
\usepackage{wrapfig}

\usepackage{comment}    
\usepackage{makecell}   

\definecolor{newcolor}{rgb}{.8,.349,.1}

\newcommand{\bx}{\bs{x}}

\newcommand{\bksi}{\bs{\xi}}

\newcommand{\by}{\bs{y}}

\newcommand{\bxx}{\bs{X}}

\newcommand{\byy}{\bs{Y}}

\newcommand{\bt}{\bs{\theta}}

\newcommand{\brr}{\mathcal{R}}

\newcommand{\beps}{\bs{\epsilon}}

\newcommand{\be}{\begin{equation}}
\newcommand{\ee}{\end{equation}}
\newcommand{\bc}{\begin{center}}
\newcommand{\ec}{\end{center}}
\newcommand{\bd}{\begin{description}}
\newcommand{\ed}{\end{description}}
\newcommand{\bi}{\begin{itemize}}
\newcommand{\ei}{\end{itemize}}
\newcommand{\pa}{\partial}
\newcommand{\bs}{\boldsymbol}

\newcommand{\bz}{\bs{z}}

\def\RR{ \mathbb R}

\newcommand{\refeqp}[1]{Equation (\ref{#1})}

\newcommand{\bmat}{\begin{pmatrix}}
\newcommand{\emat}{\end{pmatrix}}
\newcommand{\bsmat}{\left(\begin{smallmatrix}}
\newcommand{\esmat}{\end{smallmatrix}\right)}
\newcommand{\bes}{\begin{equation}\begin{split}}
\newcommand{\ees}{\end{split}\end{equation}}

\definecolor{TUMGruen}{RGB}{162,173,0} 
\definecolor{TUMOrange}{RGB}{227,114,34} 

\begin{document}

\begin{frontmatter}



\title{A probabilistic generative model for semi-supervised training of  coarse-grained  surrogates  and enforcing physical constraints through virtual observables}


\author[mymainaddress]{Maximilian Rixner}
\ead{maximilian.rixner@tum.de}

\author[mymainaddress]{Phaedon-Stelios Koutsourelakis\corref{mycorrespondingauthor}}
\cortext[mycorrespondingauthor]{Corresponding author}
\ead{p.s.koutsourelakis@tum.de}

\address[mymainaddress]{Professorship of Continuum Mechanics, Technical University of Munich}

\begin{abstract}
The data-centric construction of inexpensive surrogates for fine-grained, physical models has been at the forefront of computational physics due to its  significant utility in many-query tasks such as uncertainty quantification.   Recent efforts have taken advantage of the  enabling technologies from the field of machine learning (e.g.  deep neural networks) in combination with   simulation data. While such strategies have shown promise even in higher-dimensional problems, they generally require large amounts of training data even though the construction of  surrogates is by definition a Small Data problem.  Rather than employing data-based loss functions, it has been proposed to make use of the governing equations (in the simplest case at collocation points) in order to imbue domain knowledge in the  training of the otherwise black-box-like interpolators.  
The present paper provides a flexible, probabilistic framework that accounts for physical structure and information both in the training objectives as well as in the surrogate model itself. We advocate a {\em probabilistic} (Bayesian) model in which equalities that are available from the physics (e.g. residuals, conservation laws) can be introduced as  {\em virtual} observables and can provide additional information through the likelihood. 
We further advocate a generative model i.e. one that attempts to learn the joint density of inputs and outputs that is capable of making use of {\em unlabeled} data (i.e. only inputs) in a semi-supervised fashion in order to promote  the discovery of lower-dimensional embeddings which are nevertheless predictive of the fine-grained model's output. 
\end{abstract}



\begin{keyword}
probabilistic machine learning \sep virtual observables \sep  high-dimensional surrogates \sep semi-supervised learning \sep unlabeled data



\end{keyword}

\end{frontmatter}


\newpage
\section{Introduction}
\label{sec:intro}

The complexity and cost of many models in computational physics, necessitates the development of less expensive surrogates (or coarse-grained/reduced-order models) that provide insight but more importantly retain predictive accuracy and can enable {\em many-query} applications such as uncertainty quantification.
The difficulty of such problems is amplified in the high-dimensional setting i.e. when the number of input-output (random) variables is large as in most cases of practical interest. Data-based surrogates must also be capable of dealing with the scarcity of training data \citep{koutsourelakis_special_2016}. Unlike recent successes  in statistical/machine learning, and supervised learning in particular,  which in large part have been enabled by large datasets (and the computational means to leverage them), the acquisition of data, i.e. pairs of input-outputs, is the most expensive task and the reduction of their number, the primary objective of surrogate development.  

Another critical challenge stems from the nature of the physical models themselves. Their primary utility  arises from  their ability to distill apparent complexity and high-dimensional descriptions into much fewer, essential variables and the relations between them, which can in turn be used to make accurate predictions under a variety of settings (e.g. different boundary/initial conditions, right-hand-sides etc). This robustness of  physical models as well as their ability to operate under {\em extrapolative} conditions is not a property shared by black-box statistical surrogates, which in most cases are used in {\em interpolative} settings.

We argue that to overcome these challenges, domain knowledge i.e. information about the underlying physical/mathematical structure of the problem, must be injected into the surrogates constructed  \citep{marcus_rebooting_2019}.
 While this prior physical knowledge is generally plentiful and eloquently reflected in the governing equations, it is not necessarily obvious  neither how to to mine it nor how to automatically combine it with the data-based learning objectives, especially in a  probabilistic setting \citep{stewart2017label}. 
We believe that a probabilistic framework  provides a superior setting for such problems as it is  capable of quantifying predictive uncertainties which are unavoidable when any sort of model/dimensionality reduction is pursued and when the surrogate model is learned from finite (and hopefully, small) data \citep{koutsourelakis2007stochastic}.

The development of  surrogates for the purposes of uncertainty quantification  in the context of continuum thermodynamics where pertinent models are based on PDEs and ODEs has a long history. Some of the most well-studied methods have been based on  (generalized) Polynomial Chaos expansions (gPC) \citep{Ghanem1991, Xiu2002} which have gained  popularity  due to the emergence of data-based, non-intrusive,  sparse-grid stochastic collocation approaches \citep{Xiu2005, Ma2009, Lin2009}. These approaches typically struggle with high-dimensional stochastic inputs, as is the case  e.g. when  random heterogeneous media \citep{Torquato1993} are considered. 

Another strategy for  the construction of inexpensive surrogates is offered by reduced-basis (RB) methods \citep{Hesthaven2016, Quarteroni2016} where, based on a small set of "snapshots" i.e. input-output pairs, the solution space's dimensionality is reduced by projection onto the principal  directions. Classical formulations rely on  (Petrov-)Galerkin projections \citep{Rowley2004} for finding the associated coefficients, but recently several efforts have been directed towards unsupervised and  supervised learning  strategies \citep{Guo2018, Hesthaven2018,kani2017dr,wang2020recurrent}. Apart from issues of efficiency and stability, RB approaches in their standard form are generally treated in a non-Bayesian way and therefore only yield point estimates instead of full predictive posterior distributions.
 Furthermore, since scalar- or vector- or matrix-valued quantities need to be learned as a function of the parametric input in the  offline phase, they are also challenged by the high-dimensions/small-data setting considered \citep{lee2020model}. 

A more recent trend is to view surrogate modeling as a supervised learning problem and employ pertinent statistical learning tools, e.g. Gaussian Process (GP) regression \citep{Rasmussen2006,Bilionis2013,Bilionis2017} and which can frequently provide closed-form predictive distributions. Although several advances have been made towards multi-fidelity data fusion \citep{Kennedy2000, Koutsourelakis2009, Raissi2017, Perdikaris2015} and incorporation of  physical information \citep{Yang2018, Lee2018, Tipireddy2018b,guo2018reduced} via Gaussian Processes, their performance and  scaling  with stochastic input dimension remains one of the main challenges for GP models.  In the context of supervised learning, deep neural networks (DNNs) \citep{LeCun2015, Goodfellow2016} have found their way into surrogate modeling of complex computer codes \citep{han_solving_2018, zhu2018bayesian, Mo2018b,sirignano_dgm_2018,e_deep_2018}. One of the most promising  developments in the adaptation of such tools for physical modeling are physics-informed neural networks \citep{Raissi2017b, Raissi2018, Raissi2019, Yang2019} which are trained by minimizing  a loss function augmented by the residuals of the governing equations \citep{lagaris_neural-network_2000}.
 Physical knowledge in training DNNs has also been introduced in the form of residuals in  \citep{Raissi2017b,kani2017dr,nabian2018deep,beck_machine_2019, karumuri2020simulator, khodayi-mehr_varnet:_2019} whereas in \citep{zhu2019physics}, a Boltzmann-type  density containing physics-based  functionals or residuals were employed as the target for the associated learning problem.
 Recent reviews of the use of various machine learning models, and in particular deep neural networks, for the solution of problems in computational physics, including the development of surrogates, can be found in \citep{frank_machine-learning_2020, willard2020integrating}. Therein the difficulty of the task of   incorporating  physical, domain-knowledge  into machine learning objectives and tools \citep{mattheakis_physical_2020,magiera2020constraint}  is detailed as well as the scarcity of probabilistic approaches in the context of such tasks. 
 
 In contrast to the majority of the efforts detailed above, our goal is not to develop approximations to the PDE-solution but to capture its dependence on  high-dimensional parametric vectors. For that purpose we consider as our reference model a discretized version of the PDE which is assumed to provide sufficiently accurate resolution  (we refer to this as the Fine-Grained Model (FGM)).
 Furthermore, we wish to differentiate our work from applications of  machine learning in problems where the underlying governing equations themselves are assumed unknown and one aims to identify them from data \citep{brunton2016sparse, long2017pde, felsberger_physics-constrained_2019}. While a component of our model makes use of a (discretized) coarse-grained  model, its form is in this work prescribed.



We propose overcoming  the aforementioned challenges by introducing  a novel, generative probabilistic model that is capable of exploiting labeled (i.e. input-output pairs) and unlabeled (i.e. only inputs) data in discovering  lower-dimensional embeddings and identifying the right surrogate model-structure (section \ref{sec:method}).
More importantly, we propose augmenting the aforementioned data by injecting domain knowledge in a principled manner in the probabilistic models employed. In particular, such physical/mathematical knowledge is incorporated:
\bi
\item in the learning objectives (section \ref{sec:VirtualObservables}) through the novel notion of {\em virtual} observables \citep{kaltenbach2019incorporating}. We demonstrate how various types of information in the form of (non)linear equalities/constraints as well as minimizing functionals can be introduced in the  likelihood terms.
\item in an appropriately selected coarse-grained model  (CGM, section \ref{sec:physics_inspired_structure}) which through  coarsened or reduced-physics versions of the  full-order model provides an integral component of the proposed surrogate.  
\ei
We complement the aforementioned elements with a integrated, supervised dimensionality reduction scheme which can distill lower-dimensional features of the high-dimensional input that are most predictive of the high-dimensional output and which is trained simultaneously with the other components by making use of (un)labeled data and virtual observables.
We employ Stochastic Variational Inference procedures for training the proposed model (section \ref{sec:inference}), which upon completion yields a  probabilistic surrogate that not only produces point estimates of the high-dimensional output but can quantify the predictive uncertainty associated with this task (section \ref{sec:predictions}).
We assess the predictive performance of the proposed framework in section \ref{sec:num} where we demonstrate that unlabeled data and virtual observables can lead to significant improvements in its generalization accuracy and can  reduce the number of labeled data (i.e. input-outputs pairs) to a few tens. Furthermore, we illustrate the model's ability to [perform equally well under interpolative and extrapolative conditions i.e. under boundary conditions seen or not seen during training. We finally demonstrate its benefits in an uncertainty propagation problem and discuss possible extensions in section \ref{sec:conclusions}.


\section{Methodology}
\label{sec:method}

\vspace{1cm}

We consider steady-state physical processes modeled by a partial differential equation:
\be
\mathcal{L}(u(\bs{s}); \bx)=0, \quad \bs{s} \in \Omega
\label{eq:pde}
\ee
over the physical domain $\Omega \subset \RR^d$. The differential operator $\mathcal{L}$  depends on the random parameters $\bx \in \RR^{d_x}$ and so does the 
 solution of the PDE  $u(\bs{s})$. We denote by $\by \in \RR^{d_y}$ discretized versions of the latter and  by $\by(\bx)$ the input-output map implied by  any of the usual PDE-discretization schemes. The governing equations are complemented by boundary conditions which might partially or completely depend on the parameters $\bx$. We refer to this discretized model as {\em fine-grained model} (FGM).
 We are interested in FGMs that are computationally demanding, i.e. the number of forward model runs determines  the cost of the analysis task of interest (e.g. forward or backward uncertainty propagation, optimization). Furthermore, the problems of interest are high-dimensional, i.e.  $d_x, d_y >>1$, as in most cases of practical interest. Our goal is to construct a surrogate with the {\em least} possible {\em labeled} data $N_l$ i.e. input-output pairs $\mathcal{D}_l =\{\bx^{(i)}, \by^{(i_l)}=\by(\bx^{(i_l)}) \}_{i_l=1}^{N_l}$.  \\
 
 Several probabilistic {\em  discriminative} models i.e. models that attempt to learn $p(\by | \bx)$ have been proposed e.g. using Gaussian Processes \cite{bilionis_multi-output_2013}, Polynomial Chaos \citep{xiu2010numerical, xiu2002modeling}  or more recently using Deep Neural Networks 
 \citep{Raissi2019,zhu2018bayesian, nabian2018deep,khodayi-mehr_varnet:_2019}. It is clear that in the {\em Small Data} setting, such attempts can be generalizable only if the problem is amenable to dimensionality reductions i.e. a lower-dimensional set of features of $\bx$ are predictive of $\by$ and/or the latter itself lives in a lower-dimensional manifold. The simultaneous discovery of such lower-dimensional embeddings through a latent variable model was demonstrated in \cite{grigo_bayesian_2019,grigo_physics-aware_2019}  where  the sought density $p(\by|\bx)$ was approximated by:
 \be
 p_{\bt}(\by|\bx)= \int p_{\bt}(\by|\bs{z} )~p_{\bt}(\bs{z}| \bx)~dx
 \label{eq:discrlatent}
 \ee
 with $\bt$ being the trainable parameters of the model. The variables $\bs{z} \in \RR^{d_z}$ represent the lower-dimensional (i.e. $d_z<< d_x,d_y$) information bottleneck between inputs and outputs. In the aforementioned works, these have been associated with a lower-fidelity physical model and have been identified in the presence of Small Data using Sparse Bayesian learning  from a large vocabulary of  physically-motivated  features of $\bx$. 
 
 
\subsection{Generative Model} 
 
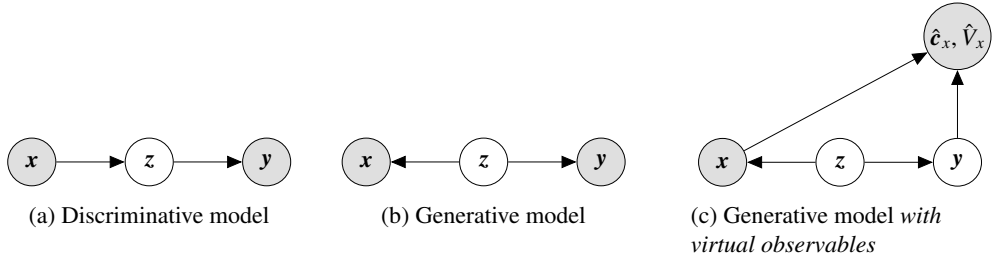
\begin{figure}
  \centering
  \begin{subfigure}{.3\linewidth}
    \centering
\scalebox{0.9}{
\begin{tikzpicture}
\node[latent] (z) {$\mat{z}$};
\node[obs, left=of z] (x) {$\mat{x}$};
\node[obs, right=of z] (y) {$\mat{y}$};
  \edge{x} {z};
  \edge{z} {y};
\end{tikzpicture}
    }
    \caption{Discriminative model}
        \label{fig:ProbabilisticModel:a} 
  \end{subfigure}%
  \hspace{1em}
  \begin{subfigure}{.3\linewidth}
    \centering
\scalebox{0.9}{
\begin{tikzpicture}
\node[latent] (z) {$\mat{z}$};
\node[obs, left=of z] (x) {$\mat{x}$};
\node[obs, right=of z] (y) {$\mat{y}$};
  \edge{z} {x};
  \edge{z} {y};
\end{tikzpicture}
}
    
    \caption{Generative model}
        \label{fig:ProbabilisticModel:b} 
  \end{subfigure}%
  \hspace{2em}
  \begin{subfigure}{.3\linewidth}
  \label{Fig:b}
    \centering
    \vspace{-1.40cm}
\scalebox{0.9}{
\begin{tikzpicture}

 
\node[latent] (z) {$\mat{z}$};
\node[obs, left=of z] (x) {$\mat{x}$};
\node[latent, right=of z] (y) {$\mat{y}$};
\node[obs, above=of y] (o) {$\hat{\bs{c}}_x, \hat{V}_x$};
  \edge{z} {x};
  \edge{z} {y};
  \edge{x}{o};
  \edge{y}{o};
\end{tikzpicture}
}
 
    \caption{Generative model \textit{with} \\ \textit{virtual observables}}
        \label{fig:ProbabilisticModel:c} 
  \end{subfigure}
  \caption{Illustration of differences between probabilistic graphical models discussed. a) {\em Discriminative} model where the latent variables $\bz$ encode lower-dimensional features of the input $\bx$ which are predictive of the output $\by$, b) {\em Generative} model where $\bz$ represent latent generators of both input and output, and c) {\em Generative} model where (b) is augmented by {\em virtual observables} encoding domain knowledge.}
\end{figure}

 The first novel contribution of this work is the use of a {\em generative }
 model i.e. one that attempts to approximate the {\em joint} density $p(\bx, \by)$ and which can subsequently be used by conditioning on $\bx$ for predictive purposes. Such a model has the  capability of  ingesting {\em unlabeled} data (i.e. only inputs) $\mathcal{D}_u=\{ \bx^{(i_u)} \}_{i_u=1}^{N_u}$ and therefore enable \textit{semi-supervised} learning. This in turn  allows the use of the information provided by the inexpensive (and potentially large) dataset $\mathcal{D}_u$ which can reduce the dependence on the expensive labeled data \cite{Chapelle2010,kingma_semi-supervised_2014}. In particular, we propose a model that performs supervised dimensionality reduction of $\bx$ and $\by$ \cite{yu_supervised_2006} and for each labeled pair $i_l$ in $\mathcal{D}_l$ assigns a likelihood:
 \be
p_{\bt} (\bx^{(i_l)},\by^{(i_l)}  )=\int p_{\bt} (\by^{(i_l)} |\bz^{(i_l)}) ~p_{\bt}(\bx^{(i_l)} |\bz^{(i_l)})~p_{\bt}(\bz^{(i_l)})~d\bz^{(i_l)}
\label{eq:likelabeled}
 \ee
 We denote again with $\bt$ any tunable model parameters  although these are in general different from the ones in \refeqp{eq:discrlatent}.
 The   unobserved variables  $\bz$  play the role of latent generators of $\bx$ and $\by$ which constitute them (conditionally) independent.
  We specify the form of the aforementioned densities, their parameterization as well as their training in the sequel. We note that the generative construction adopted  provides also a likelihood for each unlabeled datapoint $i_u$ in $\mathcal{D}_u$  as follows:
 \be
 p_{\bt}(\bx^{(i_u)})=\int p_{\bt}(\bx^{(i_u)}|\bz^{(i_u)}) p_{\bt}(\bz^{(i_u)}) ~d\bz^{(i_u)}
 \label{eq:likeunlabeled}
 \ee
 Furthermore,  for predictive purposes,  the posterior of $\bz$ for a new  $\bx$ i.e. $p_{\bt}(\bz|\bx) \propto p_{\bt}(\bx|\bz)p_{\bt}(\bz)$ can be used in order to compute:
 \be
 p_{\bt}(\by|\bx) = \int p_{\bt}(\by, \bz | \bx) ~d\bz= \int p_{\bt}(\by | \bz)p_{\bt}(\bz|\bx) ~d\bz
 \label{eq:genpred}
 \ee
 i.e. the predictive posterior on the corresponding output $\by$.
 Figures \ref{fig:ProbabilisticModel:a} and \ref{fig:ProbabilisticModel:b} provide illustrations of the discriminative and generative probabilistic graphical models.

\subsection{Virtual Observables}
\label{sec:VirtualObservables}

The second novelty proposed in this paper pertains to the introduction of domain knowledge as represented in the governing equation (\refeqp{eq:pde}) into the learning objectives. 
We would like the training process not to rely exclusively on  unlabeled $\mathcal{D}_u$ or labeled $\mathcal{D}_l$ data but rather on physical knowledge and information. These  can appear in several forms but since we are interested in their systematic incorporation we consider here various (in)equalities expressing different types of physical relations between the model-variables. The  governing PDE of \refeqp{eq:pde} for example, is a potentially {\em infinite source} of information (if one considers that the equality  holds at each of the infinite points of the problem domain $\Omega$) in contrast to the limited times these governing equations can be solved due to computational expense. While the introduction of such equalities is rather straightforward in deterministic settings in the training loss and has been employed successfully in the context of physics-informed neural networks (PINNs \cite{Raissi2019}), in a probabilistic setting,  it has only been achieved for linear ones and in order to approximate the solution of the PDE (not its dependence on input parameters) using Gaussian Processes \citep{raissi2017machine}. In this work, we generalize the type of equalities that we consider by including nonlinear ones as well as demonstrate how other types of information, e.g. that solution is a minimizer of a functional, can be introduced. We discuss below such how these can be incorporated in the learning/inference process and we give specific examples of the forms these take in the numerical illustrations (section \ref{sec:num}).

Consider first  equality constraints, i.e.:
\be
\bs{c}(\by; \bx)=\bs{0}
\label{eq:constr}
\ee
where $\bs{c}: \RR^{d_y} \times \RR^{d_x} \to \RR^{d_c}$. Such equalities can represent residuals of the governing PDE computed e.g. at some collocation points or by employing weighted residuals with appropriate test functions. They might also represent the enforcement of a physical constraint such as a conservation law (e.g. mass, momentum, energy). 
The only requirement  on $\bs{c}$ imposed by our framework is that they are {\em differentiable} functions, a property that will prove crucial in the Variational Inference component (section \ref{sec:inference}).
In order to incorporate \refeqp{eq:constr}, we introduce an auxiliary variable/vector $\hat{\bs{c}}_x$ which relates to $\bs{c}$ as follows:
\be
\hat{\bs{c}}_x=\bs{c}(\by; \bx)+\sigma_c \beps_c, \qquad \beps_c \sim \mathcal{N}(\bs{0}, \bs{I})
\label{eq:voconstr}
\ee
We further assume that $\hat{\bs{c}}_x$ is {\em virtually observed} and $\hat{\bs{c}}_x=0$. This induces a virtual likelihood $p(\hat{\bs{c}}_x| \bx,\by)$ i.e.:
\be
p(\hat{\bs{c}}_x=\bs{0}~|~ \bx,\by) \propto \frac{1}{\sigma_c^{d_c/2}} ~e^{  -\frac{1}{2 \sigma_c^2} ~|~\bs{c}(\by; \bx)~|^2 }
\label{eq:volike}
\ee
The parameter $\sigma_c$ determines the intensity of the enforcement of the virtual observation and is analogous to the tolerance parameter with which constraints or residuals are enforced in a deterministic solvers. In the limit that $\sigma_c \to 0$, the likelihood above degenerates to a Dirac-delta concentrated on the manifold implied by the constraint.
We note further that in the context of the generative model one can exploit additional unlabeled data, $\{ \bx^{(i_c)}, \hat{\bs{c}}_x^{(i_c)}\}$ consisting of pairs of inputs and {\em virtual observables} and the likelihood of each such data-pair $i_c$ will be:
\be
\begin{array}{ll}
p_{\bt}(\bx^{(i_c)}, \hat{\bs{c}}_x^{(i_c)}=\bs{0}) & = \int p_{\bt}(\hat{\bs{c}}_x^{(i_c)} ,  \by^{(i_c)}, \bz^{(i_c)}, \bx^{(i_c)}) ~d\by^{(i_c)}~d\bz^{(i_c)} \\
& = \int p(\hat{\bs{c}}_x^{(i_c)}=\bs{0}~ | \by^{(i_c)},  \bx^{(i_c)}) p_{\bt}(\by^{(i_c)}, \bz^{(i_c)}, \bx^{(i_c)}) ~d\by^{(i_c)}~d\bz^{(i_c)} \\
& = \int p(\hat{\bs{c}}_x^{(i_c)}=\bs{0}~ | \by^{(i_c)},  \bx^{(i_c)}) p_{\bt}(\by^{(i_c)} | \bz^{(i_c)}) p_{\bt}(\bx^{(i_c)} |\bz^{(i_c)}) ~p_{\bt}(\bz^{(i_c)})~d\by^{(i_c)}~d\bz^{(i_c)} 
\end{array}
\label{eq:likeconstr}
\ee
We emphasize that in this case, the solution vector $\by^{(i_c)}$ (which satisfies the constraint $\bs{c}(\by^{(i_c)}; \bx^{(i_c)})$) is latent and must be inferred. 
The corresponding graphical model is illustrated in Figure  \ref{fig:ProbabilisticModel:c} where the virtual observables represent an  observed node  \citep{levine2018reinforcement} with $\vect{y}$ - the solution of the PDE - becoming a latent variable and therefore unknown quantity in this case. \\

Another type of physical information that can be accommodated with the concept of virtual observables pertains to the variational nature of the associated problem. It is well-known that the solution of most PDEs in computational physics can be written as minimizers of an appropriate functional. 
Such functionals have served as the foundation of several numerical schemes and appear in various forms, even for irreversible, nonlinear processes \citep{ortiz_variational_1999,yang_variational_2006}. 
Various versions of these functionals were incorporated in the machine learning loss functions \cite{khoo2017solving} as well as likelihood  functions in probabilistic versions \cite{zhu2019physics}.

Suppose, that the discretized solution vector $\by(\bx)$ is obtained as the minimizer of:
\be
\by(\bx) = \arg \min_{\by} V(\by;\bx)
\label{eq:potential}
\ee
where $V: \RR^{d_y} \times \RR^{d_x} \to \RR$ represents a generalized free energy or potential. Let $V_{min}(\bx) =\min_{\by} V(\by;\bx)$ be the unknown minimum value of $V$ (attained by the solution) for each $\bx$. We define the new variable $\hat{V}_x$ as follows:
\be
\hat{V}_x= V(\by;\bx)-V_{min}(\bx)-\epsilon_V, \qquad \epsilon_V\sim \text{Expon}(\beta^{-1})
\label{eq:vopot}
\ee
The random variable $\epsilon_V$ is by construction always non-negative and follows an exponential distribution with parameter $\beta$\footnote{$\epsilon_V$ can be thought as the probabilistic analogue of a slack variable for the enforcement of inequality constraints in optimization}. We further assume that $\hat{V}_x=0$ has been {\em virtually} observed which implies a {\em virtual likelihood}:
\be
p(\hat{V}_x =0 ~| ~\by, \bx) = \beta^{-1} ~e^{ -\beta^{-1} \left( V(\by;\bx)-V_{min}(\bx) \right) }
\label{eq:likepot}
\ee
As it will be become clear in the sequel, the unknown $V_{min}(\bx)$ does not enter the training of the model. One can deduce  from Equation \eqref{eq:likepot} that the smaller $V(\by;\bx)$ is, the higher the corresponding likelihood becomes and the latter is maximized for the $\by$ that corresponds to the solution (\refeqp{eq:potential}). Furthermore, the parameter $\beta$ dictates the decay of the likelihood for $V(\by; \bx) > V_{min}(\bx)$ and in the limit $\beta^{-1} \to 0$, the likelihood degenerates to a Dirac-delta concentrated at the minimum (i.e. the true solution).

As in the previous case of the equality constraints, the introduction of these new observables enables the incorporation of the information contained in the discretized functional $V$ in the training of the proposed generative model. In particular, given  unlabeled data $\{ \bx^{(i_V)}, \hat{V}_x^{(i_V)}\}$ consisting of pairs of inputs and {\em virtual observables} $\hat{V}_x$ , the likelihood implied by the model for  each  data-pair $i_V$ will be:
\be
\begin{array}{ll}
p_{\bt}(\bx^{(i_V)}, \hat{V}_x^{(i_V)}=0) & =\int p_{\bt}( \hat{V}_x^{(i_V)}=0, ~\by^{(i_V)}, \bz^{(i_V)},\bx^{(i_V)})~d\by^{(i_V)} ~d\bz^{(i_V)} \\
& = \int p( \hat{V}_x^{(i_V)}=0~|~ \by^{(i_V)}, \bx^{(i_V)})~p_{\bt}(\by^{(i_V)}, \bz^{(i_V)}, \bx^{(i_V)}) ~d\by^{(i_V)}~d\bz^{(i_V)} \\
& = \int  p( \hat{V}_x^{(i_V)}=0~|~ \by^{(i_V)}, \bx^{(i_V)})~ p_{\bt}(\by^{(i_V)} | \bz^{(i_V)}) p_{\bt}(\bx^{(i_V)} |\bz^{(i_V)}) ~p_{\bt}(\bz^{(i_V)})~d\by^{(i_V)}~d\bz^{(i_V)} 
\end{array}
\label{eq:likepot1}
\ee
As in  \refeqp{eq:likeconstr}, the solution vector $\by^{(i_V)}$ (which minimizes $V(\by; \bx^{(i_V)}$) is latent and must be inferred.  \\

To make our notation independent of specific choices in the remainder we will introduce as a general notation a dataset of virtual observables $\mathcal{D}_{\vo} = \big\lbrace 
\vect{x}^{(i_{\vo})}, \obs^{(i_{\vo})} \big\rbrace_{i=1}^{N_{\mathcal{O}}}$, where $\vect{x}^{(i_{\vo})}$  denotes a \textit{input query point} and the corresponding $\vect{\hat{o}}^{(i_{\vo})} \in \mathbb{R}^{M}$ comprises the corresponding virtually observed values. Without loss of generality, we we assume that we enforce the same number of $M$ constraints at every point (this assumption can easily be relaxed). Parameters that govern how rigidly the constrains are enforced - such as $\sigma_c^{-1}$ or $\beta$ - are denoted summarily by  $\tau$; in the more general case, different constraints can be enforced to varying degrees, i.e. $\tau$ can comprise several precision-type parameters and may be a set instead of a scalar. We refer to each  $\vect{x}^{(i_{\vo})}$ appearing in $\mathcal{D}_{\mathcal{O}}$ as \textit{input query points} to emphasize that in the general case the corresponding solution of the PDE $\vect{y} \left( \vect{x} \right)$ is {\em not} observed/known, and we only \textit{query} certain information from the underlying physics (any equality or inequality constraint implies a certain amount of information about the underlying physics which can be exploited to inform the surrogate and to reduce epistemic uncertainty).
The introduction of virtual observables implies that the plausibility of each model contained within the hypothesis space of the generative model $p_{\bt} \left( \vect{y}, \vect{x} \right)$  is scored not only according to its performance on  unlabeled and labeled data, but also with respect to the (in)equalities that these represent.


\subsection{Physics-inspired structure for surrogate}
\label{sec:physics_inspired_structure}

The third contribution of the paper in the direction of imbuing physical knowledge into the machine learning framework pertains to the meaning of the latent variables $\bz$ and the density $p_{\bt}(\by |\bz)$. While one can make use of a purely statistical model by employing e.g. a GP or a  (deep) neural network, we advocate  here building the surrogate around a {\em coarse-grained model} (CGM). The latter can be based on simply coarsening the discretization of the governing equations (\cite{grigo_bayesian_2019}) or by employing simplified physics (\cite{grigo_physics-aware_2019}). It serves as a stencil that automatically retains the primary physical
characteristics of the FGM and enables therefore training with small amounts of data.  \\

\begin{figure}[t]
  \begin{center}
\scalebox{0.75}{
\begin{tikzpicture}

 
\node[latent] (z) {$\mat{z}$};
\node[obs, left=of z] (x) {$\mat{x}$};
\node[latent, right=of z] (X) {$\mat{X}$};
\node[latent, right=of X] (y) {$\mat{y}$};
\node[obs, above=of y] (o) {$\hat{\bs{c}}_x, \hat{V}_x$};

  \edge{z} {x};
  \edge{z} {X};
  \edge{X}{y};
  \edge{x}{o};
  \edge{y}{o};

\end{tikzpicture}
}

  \end{center}
\caption{The  node $\mat{X}$ corresponds to the (input) variables of a deterministic coarse-grained model (CGM), implying that $\vect{z}$ is encouraged not only to learn a representation of the inputs $\vect{x}$, but in particular features that through the CGM can be predictive of the FGM output $\by$ (compare with Figure \ref{fig:ProbabilisticModel:c}).}
\label{fig:pgmfull}
\end{figure}
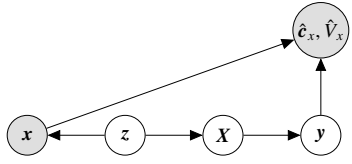

Let $\bxx$ and $\byy$ denote the input and output vector of the aforementioned CGM. The physical meaning of these variables does not need to be the same as $\bx$ or $\by$ but are by construction lower-dimensional and the solution of the CGM i.e. the cost of each evaluation of $\byy(\bxx)$\footnote{We assume a deterministic CGM for simplicity although this can be relaxed.} is negligible as compared to $\by(\bx)$.
We propose:
\bi 
\item linking the latent features $\bz$ with $\bxx$ through a density $p_{\bt} (\bxx |\bz )$ with  tunable parameters $\bt$
\item linking the sought  FGM output $\by$ with the output of the CGM $\byy(\bxx)$  rather than with $\bz$ directly. Hence instead of $p_{\bt}(\by | \bz)$ we propose employing  a density:
\be
p_{\bt} (\by~ | ~\byy(\bxx) )
\ee
\ei

The proposed model implicitly defines  an (analytically intractable) conditional  density $p_{\bt} \left( \vect{y} \middle| \vect{x} \right)$, i.e.:
\be
p_{\bt} \left( \vect{y} \middle| \vect{x} \right) = \int p_{\bt} (\by~ | ~\byy(\bxx) )~p_{\bt} (\bxx| \bz )~p_{\bt}(\bz| \bx) \diff \vect{X} \diff \vect{z}
\label{eq:predpost0}
\ee
by mapping the the latent encoding $\vect{z}$ to the \textit{effective properties} $\mat{X}$ of a CGM while simultaneously learning to reconstruct the FGM's solution $\vect{y} \left( \vect{x} \right)$ from the output of the CGM by means of $p_{\bt} \left( \vect{y} \middle| \vect{Y} \left( \vect{X} \right)  \right)$ (Figure \ref{fig:pgmfull}). 

We specify $\bxx, \byy$, the CGM itself as well as the densities involved in subsequent sections and in particular in the context of the numerical illustrations (section \ref{sec:num}). 
The introduction of the CGM and the associated latent variables $\bxx$ (and $\byy$ for a stochastic CGM) does not alter the generative nature of the model. We note though that the CGM can be omitted or simply complemented by a phenomenological statistical emulator, in which case the graphical model structure in Figure \ref{fig:pgmfull} would be altered.

\clearpage
\subsection{Specification of generative model}

\begin{figure}[t!]
    \centering
    \def\svgwidth{\columnwidth}
    \resizebox{0.70\textwidth}{!}{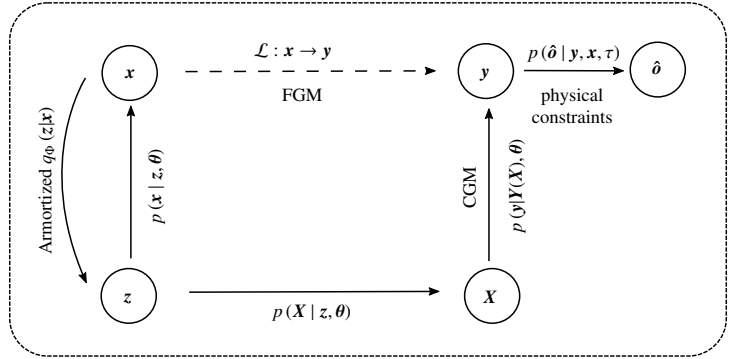}
    \caption{A schematic overview of the constituent parts of the model}
    \label{fig:architecture_overview}
\end{figure}

Based on the previous remarks and assuming real-valued $\bx, \bz,\bxx, \by$ we propose the following probabilistic generative model (for a schematic overview see also Figure \ref{fig:architecture_overview})

\begin{align}
\vect{z} &\sim \mathcal{N} \left( \vect{0}, \mat{I} \right) \label{Eq:model_0} \\
\vect{x} &= \bs{f} \left( \vect{z} ; \vect{\theta}_x \right) + \mat{S}_x^{1/2} \left( \vect{z} ; \vect{\theta}_x \right) \vect{\varepsilon}_{x}  \qquad \qquad \qquad &&\vect{\varepsilon}_x \sim \mathcal{N} \left( \vect{0}, \mat{I} \right) \label{Eq:model_A}\\
\vect{X} &= \bs{g} \left(  \vect{z} ; \vect{\theta}_g \right) + \mat{S}_X^{1/2} \vect{\varepsilon}_{X}  \qquad \qquad \qquad &&\vect{\varepsilon}_X \sim \mathcal{N} \left( \vect{0}, \mat{I} \right) \label{Eq:model_GP} \\
\vect{y} &= \bs{h} \left( \vect{Y} \left( \vect{X} \right) ; \vect{\theta}_y \right) + \mat{S}_y^{1/2}  \vect{\varepsilon}_{y} \qquad \qquad \qquad && \vect{\varepsilon}_y \sim \mathcal{N} \left( \vect{0}, \mat{I} \right) \label{Eq:model_C} 
\end{align}

where $\bs{f} \left( \cdot \right)$ and $ \bs{g} \left( \cdot \right)$ are nonlinear functions (e.g. neural networks) parameterized by $\bt_x$ and $\bt_g$ respectively. We defer  discussion of the specifics until section \ref{sec:num} where the meaning of the different variables is presented. In the following we will condition \textit{explicitly} on $\vect{\theta}$ in our notation, to avoid ambiguity and to emphasize that $\vect{\theta}$ (in the most general case) is treated probabilistically as well.
Since we operate under the assumption of {\em Small  labeled} Data, the complexity of $\bs{g} \left( \vect{z} ; \vect{\theta}_g \cdot \right)$ is chosen relatively low compared to $\bs{f} \left( \vect{z} ; \vect{\theta}_x \right)$, in order to allow learning a mapping from latent space to effective properties $\mat{X}$ with comparably few examples.
The role of $\bs{h} \left( \vect{Y} \left( \mat{X} \right); \vect{\theta}_y \right)$ is to define the map from the CGM's output $\vect{Y} \left( \vect{X} \right)$ to the (mean of the) output $\by$ of the FGM.  All the conditional densities in  \eqref{Eq:model_A} - \eqref{Eq:model_C}  are multivariate  Gaussians which have constant covariances with the exception of \refeqp{Eq:model_A}  where the covariance $\mat{S}_x$ depends on the $\bz$ variables as dictated by the associated parameters $\bt_x$.

We denote by  $\smash{\vect{\theta} = \big\lbrace \vect{\theta}_x, \vect{\theta}_g, \vect{\theta}_y, \mat{S}_x,   \mat{S}_{\vect{X}}, \mat{S}_{\vect{y}} \big\rbrace}$ the parameters of the generative model, which we wish to learn from a dataset $\smash{\mathcal{D} = \big\lbrace \mathcal{D}_{\unlabeled}, \mathcal{D}_{\labeled}, \mathcal{D}_{\vo} \big\rbrace}$ which,  in the most general case, consists of   $N_u$ unlabeled examples $\mathcal{D}_u = \big\lbrace \vect{x}^{(i_{u})} \big\rbrace_{i=1}^{N_u}$, $N_{\labeled}$ labeled input-output examples $\smash{\mathcal{D}_{\labeled} = \big\lbrace ( \vect{x}^{(i_{l})}, \vect{y}^{(i_{l})} ) \big\rbrace_{i=1}^{N_{\labeled}}}$, and tp a  collection $\smash{\mathcal{D}_{\mathcal{O}} = \big\lbrace \vect{x}^{(i_{\vo})}, \vect{\hat{o}}^{(i_{\vo})} \big\rbrace_{i_{\vo}=1}^{N_{\mathcal{O}}}}$ of query input points and virtual observables. We may then write the marginal likelihood as:

\be
\begin{array}{ll}
 p \left( \mathcal{D}  |\vect{\theta} \right) & = p(\mathcal{D}_{\unlabeled} |\bt) ~p(\mathcal{D}_{\labeled}|\bt)~p( \mathcal{D}_{\vo} |\bt) \\
 & = \prod_{i_u=1}^{N_u} p( \bx^{(i_u)} | \bt) ~ \prod_{i_u=1}^{N_l} p(\vect{x}^{(i_{l})}, \vect{y}^{(i_{l})} | \bt) ~\prod_{i_{\vo}=1}^{N_{\vo}} p(\vect{x}^{(i_{\vo})}, \vect{\hat{o}}^{(i_{\vo})} |\bt)
\end{array}
\label{eq:likedeco}
\ee
where each of the likelihood terms in the products are given by Equations (\ref{eq:likeunlabeled}), (\ref{eq:likelabeled}) and (\ref{eq:likeconstr}) (or (\ref{eq:likepot1})) respectively. In view of the densities in  Equations  \eqref{Eq:model_0} - \eqref{Eq:model_C} these become:
\be
\begin{array}{ll}
  p( \bx^{(i_u)} | \bt) = \int \mathcal{N} \left( \bx^{(i_u)} | ~\bs{f} \left( \vect{z}^{(i_u)} ; \vect{\theta}_x \right), ~\mat{S}_x \left( \vect{z}^{(i_u)} ; \theta_x \right)  \right)~  \mathcal{N} \left(\bz^{(i_u)} |~0, \bs{I} \right)~d\bz^{(i_u)},
\end{array}
\label{eq:likeu}
\ee
\be
\begin{array}{ll}
  p( \bx^{(i_l)}, \by^{(i_l)} | \bt) & = \int 
  \mathcal{N} \left(\by^{(i_l)}  |~  \bs{h} \left( \vect{Y} \left( \vect{X}^{(i_l)} \right) ; \vect{\theta}_y \right),  \mat{S}_y \right)~ 
  \mathcal{N} \left( \vect{X}^{(i_l)} |~ \bs{g} \left(  \vect{z}^{(i_l)} ; \vect{\theta}_g \right) , \mat{S}_X \right) \\
  & ~~~~~~~
  \mathcal{N} \left( \bx^{(i_l)} | ~\bs{f} \left( \vect{z}^{(i_l)} ; \vect{\theta}_x \right), ~\mat{S}_x \left( \vect{z}^{(i_l)} ; \theta_x \right)  \right)~  \mathcal{N} \left(\bz^{(i_l)} |~0, \bs{I} \right) ~d\bxx^{(i_l)}~d\bz^{(i_l)}
\end{array}
\label{eq:likel}
\ee
and
\be
\begin{array}{ll}
  p( \bx^{(i_{\vo})}, \vect{\hat{o}}^{(i_{\vo})} | \bt) & = \int 
 p( \vect{\hat{o}}^{(i_{\vo})} | \by^{(i_{\vo})}, \bx^{(i_{\vo})}; \bs{\tau}) ~ \mathcal{N} \left(\by^{(i_{\vo})}  |~  \bs{h} \left( \vect{Y} \left( \vect{X}^{(i_{\vo})} \right) ; \vect{\theta}_y \right),  \mat{S}_y \right)~ 
  \mathcal{N} \left( \vect{X}^{(i_{\vo})} |~ \bs{g} \left(  \vect{z}^{(i_{\vo})} ; \vect{\theta}_g \right) , \mat{S}_X \right) \\
  & ~~~~~~~
  \mathcal{N} \left( \bx^{(i_{\vo})} | ~\bs{f} \left( \vect{z}^{(i_l)} ; \vect{\theta}_x \right), ~\mat{S}_x \left( \vect{z}^{(i_{\vo})} ; \theta_x \right)  \right)~  \mathcal{N} \left(\bz^{(i_{\vo})} |~0, \bs{I} \right) ~d\by^{(i_{\vo})} ~d\bxx^{(i_{\vo})}~d\bz^{(i_{\vo})}
\end{array}
\label{eq:likeo}
\ee
where $p( \vect{\hat{o}}^{(i_{\vo})} | \by^{(i_{\vo})}, \bx^{(i_{\vo})}; \bs{\tau})$ depends on the nature of the virtual observable (e.g. \refeqp{eq:volike} or \refeqp{eq:likepot}). A fully Bayesian model could be defined by the introduction of appropriate priors for $\bt$ leading to to a posterior on those, i.e.  $p \left( \vect{\theta} \middle| \mathcal{D} \right) \propto p \left( \mathcal{D} \middle| \vect{\theta} \right) p \left( \bt \right)$.

\subsection{Inference and Learning}
\label{sec:inference}

Our primary objective is  to learn the model parameters  $\vect{\theta}$ on the basis of the mixed data $\mathcal{D} = \left\lbrace \mathcal{D}_u, \mathcal{D}_s, \mathcal{D}_{\mathcal{O}} \right\rbrace$ so that the trained probabilistic surrogate can be used for predictive purposes. This task is hindered by the intractability of all the likelihood terms in Equations (\ref{eq:likel})-(\ref{eq:likeo}) due to the presence of the latent variables which must be integrated-out.  
In particular, we denote summarily by $\mathcal{R}=\left\lbrace \mathcal{Z}_u,\mathcal{Z}_l,\mathcal{Z}_{\vo},\mathfrak{X}_l,\mathfrak{X}_{\vo},\mathcal{Y}_{\mathcal{O}} \right\rbrace$ the latent variables appearing in Equations (\ref{eq:likeu}) - (\ref{eq:likeo})  which consist of:
\bi
\item $\mathcal{Z}_u=\{ \bz^{(i_u)}\}_{i_u=1}^{N_u}$ associated with $\mathcal{D}_u$ (see e.g. Equation \eqref{eq:likeunlabeled} or Equation \eqref{eq:likeu}), 
\item $\mathcal{Z}_l=\{ \bz^{(i_l)}\}_{i_l=1}^{N_l}$, $\mathfrak{X}_l=\{ \bxx^{(i_l)}\}_{i_l=1}^{N_l}$ associated with $\mathcal{D}_l$ (see e.g. Equation \eqref{eq:likelabeled} or \eqref{eq:likel} ),  
\item $\mathcal{Z}_{\vo}=\{ \bz^{(i_{\vo})}\}_{i_{\vo}=1}^{N_{\vo}}$, $\mathfrak{X}_{\vo}=\{ \bxx^{(i_{\vo})}\}_{i_{\vo}=1}^{N_{\vo}}$,  $\mathcal{Y}_{\mathcal{O}}=\{ \by^{(i_{\vo})}\}_{i_{\vo}=1}^{N_{\vo}}$ associated with $\mathcal{D}_{\vo}$ (see e.g. Equation \eqref{eq:likeo}).
\ei

We advocate  the use of Stochastic Variational Inference (SVI, \citep{Paisley2012,Hoffman2013})  which produces closed-form approximations of the true posterior $p(\bt,\brr| \mathcal D)$ and simultaneously of the model evidence $p(\mathcal D)$. In contrast to sampling-based procedures (e.g. MCMC, SMC), stochastic variational inference yields biased estimates at the benefit of computational efficiency and computable convergence objectives in the form of the Evidence Lower Bound (ELBO,\citep{blei2017variational}). In particular, we denote the variational approximation to the joint posterior as $q_{\bksi}(\bt, \brr)$ where $\bksi$ are its tunable parameters and note that the model evidence $p(\mathcal{D})$ can be lower-bounded as \citep{bishop_pattern_2006}:
\be
\begin{array}{ll}
\log p(\mathcal{D} ) & = \log \int p(\mathcal{D}, \bt, \brr) ~ \diff \bt~\diff \brr \\
& = \mathcal{F} \left( \bksi \right) + KL \left( q_{\bksi} \left( \bt, \brr \right) ~\middle| \middle| ~p \left( \bt,\brr| \mathcal{D} \right) \right) \\
& \ge \mathcal{F}(\bksi)
\end{array}
\ee
where:
\be
0 \le KL \left( q_{\bksi} \left( \bt, \brr \right) ~\middle| \middle| ~p \left( \bt,\brr| \mathcal D \right) \right) = - \int q_{\bksi} \left( \bt, \brr \right) \log \left( \frac{p(\bt,\brr| \mathcal{D}) }{q_{\bksi}(\bt, \brr)} \right) ~d\bt~d\brr
\label{eq:kl}
\ee
is the KL-divergence between approximate and true posterior, and $\mathcal{F}(\bksi)$ is the ELBO i.e.
\be
\begin{array}{ll}
\mathcal{F}(\bksi) & = \int q_{\bksi} \left( \bt, \brr \right) ~\log \cfrac{ p( \mathcal{D} , \bt, \brr)}{ q_{\bksi}(\bt, \brr)}  ~d\bt~d\brr \\
& = \mathbb{E}_{q_{\bksi}} \left[ \log \cfrac{ p( \mathcal{D} , \bt, \brr)}{ q_{\bksi}(\bt, \brr)} \right] 
\end{array}
\label{eq:elbo}
\ee

Maximizing the ELBO over the parameters $\bksi$ is therefore equivalent to minimizing the KL-divergence from the true posterior. The ELBO provides a score function for comparing different approximations (e.g. different family of distributions $q \in \mathcal{Q}$ or different parametrizations $\bksi$) and as an approximation to the model evidence can also be used to compare different models (e.g. with different structure or different parametrizations $\bt$).

We employ a (partial) mean field   approximation i.e. a $q_{\bksi}$ that factorizes as follows:

\be
\begin{array}{ll}
q_{\vect{\xi}} \left( \vect{\theta}, \brr \right) & =
 q_{\vect{\xi}} \left( \vect{\theta} \right) 
  \prod_{i_u=1}^{N_{\unlabeled}} q_{\vect{\xi}} \left( \vect{z}^{(i_{\unlabeled})} \right) 
 \prod_{i_{\labeled}=1}^{N_{\labeled}} q_{\vect{\xi}} \left( \vect{z}^{(i_{\labeled})} \right) q_{\vect{\xi}} \left( \vect{X}^{(i_{\labeled})} \right)
 \prod_{i_\vo}^{N_{\vo}} q_{\vect{\xi}} \left( \vect{z}^{(i_{\vo}}\right) q_{\vect{\xi}} \left( \vect{X}^{(i_{\vo})} \right) q_{\vect{\xi}} \left( \vect{y}^{(i_{\vo})} \right).
\end{array}
\label{Eq:VariationalFactorization}
\ee
While this might appear drastic, we note that the elements of $\mathcal{Z}_u$ are conditionally (given $\bt$) independent of the rest even in the true posterior,  as also the latent variables in the following two groups $\{\mathcal{Z}_l,\mathfrak{X}_l\}$ and $\{ \mathcal{Z}_{\vo}, \mathfrak{X}_{\vo},\mathcal{Y}_{\mathcal{O}} \}$.
Given this, the ELBO becomes:

\begin{align}
  \phantom{\mathcal{F \left( \vect{\xi} \right)}}  
  &\begin{aligned}
    \mathllap{\mathcal{F} \left( \vect{\xi} \right)} &= \mathbb{E}_{q_{\vect{\xi}}} \left[ \log \left( \frac{p \left( \mathcal{D}, \vect{\theta}, \mathcal{R} \right)}{q_{\vect{\xi}} \left( \vect{\theta} , \mathcal{R} \right)} \right) \right] \nonumber
  \end{aligned}\\[6pt]
    &\begin{aligned}
    &= \mathbb{E}_{q_{\vect{\xi}}} \left[ 
    \log p \left( \mathcal{D}_u \middle| \vect{\theta}, \mathcal{R} \right) + \log p \left( \mathcal{D}_l \middle| \vect{\theta}, \mathcal{R} \right) + \log p \left( \mathcal{D}_{\vo} \middle| \vect{\theta}, \mathcal{R} \right) + \log p \left( \mathcal{R} , \vect{\theta} \right) - \log q_{\vect{\xi}} \left( \vect{\theta}, \mathcal{R} \right) 
    \right] \nonumber
  \end{aligned}\\[6pt]
  &\left. \begin{aligned}
     = &\phantom{+} \sum\nolimits_{i_u=1}^{N_u} \mathbb{E}_{q_{\vect{\xi}}} \left[ \log p \left( \vect{x}^{(i_u)} \middle| \vect{z}^{(i_u)}, \vect{\theta} \right) \right] \hphantom{+++++++++++++++} \\
      &+ \sum\nolimits_{i_l=1}^{N_l} \mathbb{E}_{q_{\vect{\xi}}} \left[ \log p \left( \vect{y}^{(i_l)} \middle| \vect{X}^{(i_l)}, \vect{\theta} \right) + \log p \left( \vect{x}^{(i_l)} \middle| \vect{z}^{(i_l)}, \vect{\theta} \right) \right]  \\
      &+ \sum\nolimits_{i_\vo=1}^{N_\vo} \mathbb{E}_{q_{\vect{\xi}}} \left[ \log p \left( \vect{\hat{o}}^{(i_\vo)} \middle| \vect{y}^{(i_\vo)}, \vect{x}^{(i_\vo)}, \vect{\theta} \right) + \log p \left( \vect{x}^{(i_\vo)} \middle| \vect{z}^{(i_\vo)}, \vect{\theta} \right) \right] 
  \end{aligned} \right\rbrace \nonumber
  \begin{aligned}
  \vphantom{ \sum\nolimits_{i_u=1}^{N_u} \mathbb{E}_{q_{\vect{\xi}}} \left[ \log p \left( \vect{x}^{(i_u)} \middle| \vect{z}^{(i_u)}, \vect{\theta} \right) \right]} \mathbb{E}_{q_{\vect{\xi}}} \left[ \log p \left( \mathcal{D}_u \middle| \vect{\theta}, \mathcal{R} \right) \right] \\
  \vphantom{ \sum\nolimits_{i_u=1}^{N_u} \mathbb{E}_{q_{\vect{\xi}}} \left[ \log p \left( \vect{x}^{(i_u)} \middle| \vect{z}^{(i_u)}, \vect{\theta} \right) \right]} \mathbb{E}_{q_{\vect{\xi}}} \left[ \log p \left( \mathcal{D}_l \middle| \vect{\theta}, \mathcal{R} \right) \right] \\
 \vphantom{ \sum\nolimits_{i_u=1}^{N_u} \mathbb{E}_{q_{\vect{\xi}}} \left[ \log p \left( \vect{x}^{(i_u)} \middle| \vect{z}^{(i_u)}, \vect{\theta} \right) \right]}  \mathbb{E}_{q_{\vect{\xi}}} \left[ \log p \left( \mathcal{D}_\vo \middle| \vect{\theta}, \mathcal{R} \right) \right]
  \end{aligned} \\[6pt]
      &\left. \begin{aligned}
      \phantom{=} &+  \sum\nolimits_{i_u=1}^{N_u} \mathbb{E}_{q_{\vect{\xi}}} \left[ \log p \left( \vect{z}^{(i_u)} \right) \right]  \\
      &+ \sum\nolimits_{i_l=1}^{N_l} \mathbb{E}_{q_{\vect{\xi}}} \left[ \log p \left( \vect{X}^{(i_l)} \middle| \vect{z}^{(i_l)}, \vect{\theta} \right) + \log p \left( \vect{z}^{(i)} \right) \right] \\
      &+ \sum\nolimits_{i_\vo=1}^{N_\vo} \mathbb{E}_{q_{\vect{\xi}}} \left[ \log p \left( \vect{y}^{(i_\vo)} \middle| \vect{X}^{(i_\vo)}, \vect{\theta} \right) + \log p \left( \vect{X}^{(i_\vo)} \middle| \vect{z}^{(i_\vo)}, \vect{\theta} \right) + \log p \left( \vect{z}^{(i_\vo)} \right) \right]
  &\end{aligned} \right\rbrace \mathbb{E}_{q_{\vect{\xi}}} \left[ \log p \left( \mathcal{R} \middle| \vect{\theta} \right) \right]  \nonumber\\[6pt]
      & \left. \begin{aligned}
      \phantom{=} &+  \mathbb{E}_{q_{\vect{\xi}}} \left[ \log p \left( \vect{\theta} \right)  \right] \\
      &- \mathbb{E}_{q_{\vect{\xi}}} \left[ \log q_{\vect{\xi}} \left( \mathcal{R} \right) + \log q_{\vect{\xi}} \left( \vect{\theta} \right) \right].
  &\end{aligned} \right. 
  \label{Eq:FullElboExpression}
\end{align}

In all subsequent illustrations we used point estimates for the parameters $\vect{\theta}$ i.e. computed their maximum-a-posteriori (MAP) estimate $\bt_{MAP}$. This is equivalent to introducing a Dirac-delta:
\be
q_{\vect{\xi}} \left( \vect{\theta} \right) = \delta \left( \vect{\theta} - \bt_{MAP} \right)
\label{eq:qksitheta}
\ee
in the variational approximation in which case the parameters $\bksi$ include also $\bt_{MAP}$. In this case, the expectations with respect to $q_{\vect{\xi}} \left( \vect{\theta} \right)$ can simply be  computed by substituting $\bt_{MAP}$ wherever $\bt$ appears and the entropy term $\mathbb{E}_{q_{\bksi}} \left[\log q_{\bksi}(\bt) \right]$ can be ignored as it is independent of $\theta_{MAP}$.

The presence of three sets of conditionally independent datasets i.e. $\mathcal{D}_u$,$\mathcal{D}_l$ and $\mathcal{D}_{\vo}$ (\refeqp{eq:likedeco}) leads to an additive decomposition of the ELBO of the form $\mathcal{F} = \mathcal{F}_{\unlabeled} + \mathcal{F}_{\labeled} + \mathcal{F}_{\vo} + \log p(\bt_{MAP})$, where
\be
\begin{array}{ll}
\mathcal{F}_u(\bksi) & = \sum_{i_u=1}^{N_u} \mathbb{E}_{q_{\bksi}} \left[  p( \bx^{(i_u)} | \bz^{(i_u)}, \bt)\right]+ \sum_{i_u=1}^{N_u} \mathbb{E}_{q_{\bksi}} \left[\log p(\bz^{(i_u)})\right]-\sum_{i_u=1}^{N_u} \mathbb{E}_{q_{\bksi}} \left[ \log q_{\bksi}(\bz^{(i_u)}) \right]
\end{array}
\label{Eq:ELBO_unsupervised} 
\ee
 accounts for the terms associated with the unlabeled data $\mathcal{D}_u$, 
 \be
\begin{array}{ll}
\mathcal{F}_l(\bksi) & =\sum_{i_l=1}^{N_l} \mathbb{E}_{q_{\bksi}} \left[ \log  p( \by^{(i_l)} | \bxx^{(i_l)}, \bt) + \log p( \bx^{(i_l)} | \bz^{(i_l)}, \bt) \right]  \\
& + \sum_{i_l=1}^{N_l} \mathbb{E}_{q_{\bksi}} \left[ \log  p(\bxx^{(i_l)} | \bz^{(i_l)}, \bt)+\log p(\bz^{(i_l)}) \right] \\
&- \sum_{i_l=1}^{N_l} \mathbb{E}_{q_{\bksi}} \left[ \log q_{\bksi}(\bxx^{(i_l)}) + \log q_{\bksi}(\bz^{(i_l)}) \right] 
\end{array}
\label{Eq:ELBO_supervised} 
\ee
 accounts for the terms associated with the labeled data $\mathcal{D}_l$, and
\be
\begin{array}{ll}
\mathcal{F}_{\vo}(\bksi) & =\sum_{i_{\vo}=1}^{N_{\vo}} \mathbb{E}_{q_{\bksi}} \left[ \log p(\vect{\hat{o}}^{(i_{\vo})} | \by^{(i_{\vo})}, \bx^{(i_{\vo})},  \bt) +\log p(\bx^{(i_{\vo})} | \bz^{(i_{\vo})}, \bt)
\right] \\
& + \sum_{i_{\vo}=1}^{N_{\vo}} \mathbb{E}_{q_{\bksi}} \left[ 
\log p(\by^{(i_{\vo})} | \bxx^{(i_{\vo})}, \bt) +\log  p(\bxx^{(i_{\vo})} | \bz^{(i_{\vo})}, \bt) +\log p(\bz^{(i_{\vo})}) \right]\\
&- \sum_{i_l=1}^{N_{\vo}} \mathbb{E}_{q_{\bksi}} \left[ \log q_{\bksi}(\by^{(i_l)})+\log q_{\bksi}(\bxx^{(i_l)}) + \log q_{\bksi}(\bz^{(i_l)}) \right] 
\end{array}
 \label{Eq:ELBO_vo}
 \ee
 accounts for the terms associated with the virtual observables/data $\mathcal{D}_{\vo}$.

We note that in \refeqp{Eq:ELBO_unsupervised}, \refeqp{Eq:ELBO_supervised} and \refeqp{Eq:ELBO_vo} the expected log-likelihood terms (i.e. first sum) promote a good fit of  the generative model to  the unlabeled $\mathcal{D}_\unlabeled$, labeled $\mathcal{D}_\labeled$ and virtual data $\mathcal{D}_{vo}$ data respectively, while the  second and third sums correspond to the Kullback-Leibler divergence between approximate posteriors and  priors  which act as regularization that prevents overfitting.
The common model parameters $\bt$ appear in all components of the ELBO and synthesize the information provided by the different data-types. We highlight the term $\log p(\vect{\hat{o}}^{(i_{\vo})} | \by^{(i_{\vo})}, \bx^{(i_{\vo})},  \bt)$ in \refeqp{Eq:ELBO_vo}, which is is driven by the virtual dataset and reflects the incorporation of our (in)equality constraints. In this case, the model attempts to  infer the solution $\by^{(i_{\vo})}$ through  $q_{\bksi} \left( \by^{(i_{\vo})}  \right)$. Hence the updates of the model parameters $\bt$ are affected also by the inferred solutions and the uncertainty associated with them.

For the structured mean-field approximation $q_{\vect{\xi}} \left( \vect{\theta}, \mathcal{R} \right)$ in Equation \eqref{Eq:VariationalFactorization} we adopt diagonal Gaussians, primarily due to their linear scaling with the dimension of the corresponding latent variables.
The following forms and parametrizations for the variational posteriors $q_{\bksi}$ in \refeqp{Eq:VariationalFactorization} were adopted:

{
\small
\renewcommand{\arraystretch}{2.1}
\begin{center}
\begin{tabular}{ l l l } 
  \textbullet $~~ \forall i_{\unlabeled} \in \left\lbrace 1, ..., N_\unlabeled \right\rbrace:$ & $q_{\vect{\xi}} \left( \vect{z}^{(i_\unlabeled)} \right) = \mathcal{N} \left( \vect{z}^{(i_\labeled)} ~ \middle| ~ \vect{\mu}_{\vect{z}}^{(i_\unlabeled)}, \text{diag} \left( \vect{\sigma}_{\vect{z}}^{(i_\unlabeled)}  \right) \right) $ &  \\ 
 \vspace{0.15cm} \textbullet $~~ \forall i_{\labeled} \in \left\lbrace 1, ..., N_\labeled \right\rbrace:$ & $q_{\vect{\xi}} \left( \vect{z}^{(i_\unlabeled)} \right) = \mathcal{N} \left( \vect{z}^{(i_\labeled)} ~ \middle| ~ \vect{\mu}_{\vect{z}}^{(i_\labeled)}, \text{diag} \left( \vect{\sigma}_{\vect{z}}^{(i_\labeled)}  \right) \right) $ & $q_{\vect{\xi}} \left( \vect{X}^{(i_\labeled)} \right) = \mathcal{N} \left( \vect{X}^{(i_\labeled)} ~ \middle| ~ \vect{\mu}_{\vect{X}}^{(i_\labeled)}, \text{diag} \left( \vect{\sigma}_{\vect{X}}^{(i_\labeled)}  \right) \right) $ \\
  \vspace{0.15cm}  \textbullet $~~ \forall i_{\vo} \in \left\lbrace 1, ..., N_\vo \right\rbrace:$ & $q_{\vect{\xi}} \left( \vect{z}^{(i_\vo)} \right) = \mathcal{N} \left( \vect{z}^{(i_\vo)} ~ \middle| ~ \vect{\mu}_{\vect{z}}^{(i_\vo)}, \text{diag} \left( \vect{\sigma}_{\vect{z}}^{(i_\vo)}  \right) \right) $ & $q_{\vect{\xi}} \left( \vect{X}^{(i_\vo)} \right) = \mathcal{N} \left( \vect{X}^{(i_\vo)} ~ \middle| ~ \vect{\mu}_{\vect{X}}^{(i_\vo)}, \text{diag} \left( \vect{\sigma}_{\vect{X}}^{(i_\vo)}  \right) \right) $ \\ 
  & $q_{\vect{\xi}} \left( \vect{y}^{(i_\vo)} \right) = \mathcal{N} \left( \vect{y}^{(i_\vo)} ~ \middle| ~ \vect{\mu}_{\vect{y}}^{(i_\vo)}, \text{diag} \left( \vect{\sigma}_{\vect{y}}^{(i_\vo)}  \right) \right) $ & 
\end{tabular}
\end{center}
}

which, in combination with \refeqp{eq:qksitheta} suggest that the parameter vector $\bksi$ consists of: 

\begin{align}
    \vect{\xi} =  \left\lbrace \vect{\theta}_{MAP}, ~ \left\lbrace \vect{\mu}_{\vect{z}}^{(i_\unlabeled)}, \vect{\sigma}_{\vect{z}}^{(i_\unlabeled)} \right\rbrace_{i=1}^{N_\unlabeled}  
    ~ \left\lbrace \vect{\mu}_{\vect{z}}^{(i_\labeled)}, \vect{\sigma}_{\vect{z}}^{(i_\labeled)}, \vect{\mu}_{\vect{X}}^{(i_\labeled)}, \vect{\sigma}_{\vect{X}}^{(i_\labeled)} \right\rbrace_{i=1}^{N_\labeled} 
    \left\lbrace \vect{\mu}_{\vect{z}}^{(i_\vo)}, \vect{\sigma}_{\vect{z}}^{(i_\vo)}, \vect{\mu}_{\vect{X}}^{(i_\vo)}, \vect{\sigma}_{\vect{X}}^{(i_\vo)},
    \vect{\mu}_{\vect{y}}^{(i_\vo)}, 
    \vect{\sigma}_{\vect{y}}^{(i_\vo)}
    \right\rbrace_{i=1}^{N_\vo}
    \right\rbrace.
    \label{Eq:FullSpecificationVariationalParameters}
\end{align}
For the parameters that are constrained to be positive, a suitable transformation (e.g. $\exp \left( \cdot \right)$) is employed such that maximizing the ELBO becomes an unconstrained optimization problem\footnote{ We note that $\vect{\sigma}$ denotes a vector of \textit{variances}, not standard deviations.}. 

We note further that, since we target cases where  that the number of labeled data points $N_l$ is as small as possible due to the numerical cost of the forward problem, one could potentially  leverage the larger number of $N_\unlabeled$ \textit{unlabeled} data which carry negligible cost. 
From \refeqp{Eq:FullSpecificationVariationalParameters} it is obvious that the number of  variational parameters with $\mathcal{D}_u$  scales linearly with $N_\unlabeled$. One may therefore consider introducing an \textit{amortized} encoder $\smash{q_{\Phi} \big( \vect{z}^{(i_{u})}  \big| \vect{x}^{(i_{u})}  \big)}$ \citep{kingma2013auto}, i.e. an approximate posterior that explicitly accounts for the dependence of each $\vect{z}^{(i_{u})}$ on the data $\vect{x}^{(i_{u})}$.
In particular, we adopt an approximate posterior of the form:
\begin{align}
q_\Phi \left( \vect{z}^{(i_\unlabeled)} \middle| \vect{x}^{(i_\unlabeled)} \right)  = \mathcal{N} \left(
\vect{z}^{(i_\unlabeled)} \middle| \vect{\mu}_\Phi \left( \vect{x}^{(i_\unlabeled)} \right), \text{diag} \left( \vect{\sigma}_\Phi \left( \vect{x}^{(i_\unlabeled)} \right) \right)
\right) \qquad \qquad  \forall i \in \left\lbrace 1, ..., N_\unlabeled \right\rbrace
\label{eq:ampostz}
\end{align}
where the parameters $\Phi$ are the same for all $i_u$. While the approximate posterior in \refeqp{eq:ampostz} can, at best, achieve the same ELBO as the  $q_{\vect{\xi}} \left( \vect{z}^{(i_\unlabeled)} \right)$ above, it contains fewer parameters that need to be optimized (at least for large $N_u$) and once trained can be readily used as an approximation to the true posterior $p_{\bt}(\bz|\bx)$ for predictive purposes in \refeqp{eq:predpost0}. In our simulations, the parameters $\Phi$ pertain to deep neural nets (see section \ref{sec:num}) and from a practical point of view, the only difference is that  $ \left\lbrace \vect{\mu}_{\vect{z}}^{(i_\unlabeled)}, \vect{\sigma}_{\vect{z}}^{(i_\unlabeled)} \right\rbrace_{i=1}^{N_\unlabeled}$  are substituted by the parameters $\Phi$ in the vector $\bksi$ of \refeqp{Eq:FullSpecificationVariationalParameters}, and that the unlabeled data is subsampled in batches during training.

\begin{algorithm}[t]
\setalgorithmspacing
    \SetAlgoLined
    \KwData{Generative Model, $\mathcal{D}_{\unlabeled} = \left\lbrace \vect{x}^{(i_\unlabeled)} \right\rbrace_{i=1}^{N_\unlabeled}$, $\mathcal{D}_{\labeled} = \left\lbrace \vect{x}^{(i_\labeled)}, \vect{y}^{(i_\labeled)} \right\rbrace_{n=1}^{N_\labeled}$ , $\mathcal{D}_{\vo} = \left\lbrace \vect{x}^{(i_\vo)}, \vect{\hat{o}}^{(i_\vo)} \right\rbrace_{n=1}^{N_{\vo}} $} 
    \While{ELBO not converged}{
        	       \tcp{Reparametrization trick}
        Sample $\vect{\epsilon}_{(k)} \sim p \left( \vect{\epsilon} \right), ~ \qquad \qquad \qquad \qquad k=1, ..., K$ \; 
        $ \mathcal{R}_{(k)}  \leftarrow \varrho_{\vect{\xi}}^{\mathcal{R}} \left( \vect{\epsilon}_{(k)} \right) \qquad  \vect{\theta}_{(k)}  \leftarrow \varrho_{\vect{\xi}}^{\vect{\theta}} \left( \vect{\epsilon}_{(k)} \right) \qquad k = 1, ..., K$  \;
        	       \tcp{Monte Carlo estimate of ELBO}
        Estimate $\mathcal{\hat{F}} \leftarrow \sum_{k=1}^K \mathcal{F} \left( \vect{\theta}_{(k)}, \mathcal{R}_{(k)} \right)$ \tcp*[r]{Equation \eqref{Eq:FullElboExpression}}  \
        \tcp{Backpropagate}
         $\vect{g}_{\vect{\xi}} \leftarrow \nabla_{\vect{\xi}} \sum_{k=1}^K \mathcal{F} \left( \vect{\theta}_{(k)}, \mathcal{R}_{(k)} \right)$ \;
        \tcp{Stochastic Gradient Update} $\vect{\xi}^{(n+1)} \leftarrow \vect{\xi}^{(n)} + \bm{\rho}^{(n)} \odot \vect{g}_{\vect{\xi}}$ \;
        $n \leftarrow n + 1$

    }
\caption{Training generative model using SVI}
\label{Alg:GeneralAlg}
\end{algorithm}

We conclude this section by enumerating the basic  steps associated with the variational inference task in Algorithm \ref{Alg:GeneralAlg}. The intractable expectations with respect to $q_{\bksi}$ appearing in the ELBO $\mathcal{F}$ and its gradient $\nabla_{\bksi} \mathcal{F}$ are estimated with Monte Carlo. In order to  reduce the variance of these estimators, we apply the well-established reparametrization trick \citep{kingma2013auto}. 

We combine the  noisy estimates of the gradient $\nabla_{\bksi} \mathcal{F}$ with  stochastic gradient ascent \citep{robbins1951stochastic} and the Adam algorithm in particular \citep{kingma2014adam}. We note that  training requires the propagation of  gradients through the whole model, including the  CGM and the constraints associated with virtual observables. Propagating gradients through the model can readily be done using algorithmic differentiation \citep{naumann2012art} whenever possible; i.e. when evaluating a Monte Carlo estimate of the evidence lower bound $\mathcal{F}$ a computational graph is built, such that in a backward pass gradient information propagates from $\mathcal{F}$ to the leaf nodes of the computational graph (e.g. given by the variational parameters $\vect{\xi}$) \citep{paszke2017automatic}. The CGM and the virtual observables $\vect{o} \left( \vect{y} ; \vect{x} \right)$ must be embedded within this computational graph, i.e require that the CGM  also allows the back-propagation of gradient information. If the CGM is also given by differential equation, the reverse-flow of information required during back-propagation corresponds to the solution of the adjoint problem, at a cost equivalent to the forward solution of the CGM. Obtaining derivatives of the virtual observables is equally a cheap operation but also problem-specific and discussion is deferred until section \ref{sec:Example_VO}.


\subsection{Predictions}
\label{sec:predictions}
Given the (approximate) posterior $q_{\bksi}(\bt)$ on the model parameters $\bt$ obtained after training, the model can be employed for predictive purposes. 
In the simplest case, given a new (unobserved) input ${\bx}$, we seek the corresponding output ${\by}$. The probabilistic nature of the proposed generative model yields a probability density on $\by$ (see also \refeqp{eq:genpred}) i.e. the predictive posterior $p \left( \by \middle| \bx, \mathcal{D} \right)$ given by 
\begin{align}
p \left( \by \middle| \bx, \mathcal{D} \right) & = \int p \left(\by \middle| \mat{X}, \vect{\theta}  \right) p \left( \mat{X} \middle| \vect{z}, \vect{\theta} \right) p \left( \vect{z} \middle| \bx, \vect{\theta} \right) p \left( \vect{\theta} \middle|  \mathcal{D} \right) \diff \vect{z} \diff \mat{X} \diff \vect{\theta} \\
&\approx \int p \left( \vect{\tilde{y}} \middle| \mat{X}, \thetamap \right) p \left( \vect{X} \middle| \vect{z}, \thetamap \right) p \left( \vect{z} \middle| \bx, \thetamap \right) \diff \mat{X} \diff \vect{z}
\label{Eq:PostPred_approx}
\end{align}
where the variational approximation $q_{\bksi}(\bt) = \delta \left( \vect{\theta} - \theta_{MAP} \right)$ was used in place of the intractable posterior $p \left( \vect{\theta} \middle| \mathcal{D} \right)$. We also briefly mention the possibility (without pursuing it further in this work) to incorporate (additional) constraints $\vect{{o}}(\by;\bx)$ at ${\vect{x}}$ during the prediction stage as well, i.e. to perform \textit{prediction by inference} and update the posterior predictive using again the \textit{virtual likelihood}

\begin{align}
    p \left( \vect{y} \middle| \vect{x}, \hat{\vect{o}}, \mathcal{D} \right) \propto p \left( \hat{\vect{o}} \middle| \vect{y}, \vect{x} \right) p \left( \vect{y} \middle| \bx, \mathcal{D} \right)
\end{align}
where $\vect{\hat{o}}$ denotes the associated virtual observables. 

 If an amortized approximate posterior $q_{\Phi}(\bz | \bx)$ has been found in the inference step as detailed in the previous section, then this can be used in place of $p \left( \vect{z} \middle| \bx, \thetamap \right)$  in Equation \eqref{Eq:PostPred_approx}. 
Alternatively, one might employ sampling methods e.g. MCMC or another round of (stochastic) variational inference in order to obtain an approximation, say  $q_{\bs{\zeta}} \left( \vect{z} \right)$. The latter is found by maximizing an analogous ELBO, i.e.:

\begin{align}
q^* \left( \vect{z} \right) &= \arg \min_{\vect{\zeta}} \text{KL} \left[ q_{\vect{\zeta}} \left( \vect{z} \right) \middle| \middle| p \left( \vect{z} \middle| \vect{x}, \thetamap \right) \right] \nonumber \\
&= \arg\max_{\vect{\zeta}} \mathbb{E}_{q_{\vect{\zeta}} \left( \vect{z} \right)} \left[
\log p \left( \bx \middle| \vect{z}, \thetamap \right)
\right] 
- \text{KL} \left[
q_{\vect{\zeta}} \left( \vect{z} \right) \middle| \middle| p \left( \vect{z} \right) \right] \nonumber \\ 
&= \arg\max_{\vect{\zeta}} \hat{\mathcal{F}}_{u} \left( q_{\vect{\zeta}} \left( \vect{z} \right) \right) \label{eq:predpost}
\end{align}
We note that in neither of the latter two cases, any additional model solves are required and in the results reported in subsequent sections the variational approximation $q_{\bs{\zeta}}$ was used. The  integral in the predictive posterior  of  \eqref{Eq:PostPred_approx} can be approximated with Monte Carlo and requires solely solutions of the  CGM. 
In Algorithm \ref{Alg:GeneralAlgPredictions} we briefly summarize how probabilistic predictions $p \left( \vect{y} \middle| {\vect{x}} , \mathcal{D} \right)$ can be obtained for new (unobserved) inputs  ${\vect{x}}$.

\begin{algorithm}[ht]
    \setalgorithmspacing
    \SetAlgoLined
    \KwData{$\bx$, trained generative model}

        \eIf{amortization}{
       		$q^* \left( \vect{z} \right) \leftarrow q_{\Phi} \left( \vect{z} \middle| \bx \right)$  \tcp*[r]{Equation \eqref{eq:ampostz}} 
            }{
        $q^* \left( \vect{z} \right) \leftarrow \arg \max_{\vect{\zeta}} \hat{\mathcal{F}}_u \left( q_{\vect{\zeta}} \left( \vect{z} \right) \right)$  \tcp*[r]{Equation \eqref{eq:predpost}} 
        }
        
	\For{$k \leftarrow 1$ \KwTo $K$}{
	   Sample $\vect{z}^{(k)} \sim q^* \left( \vect{z} \right) $ \;
	   Sample $\vect{X}^{(k)} \sim p \left( \vect{X} \middle| \vect{z}^{(k)}, \thetamap \right)$  \tcp*[r]{Equation \eqref{Eq:model_GP}}  \
       Sample $\vect{y}^{(k)} \sim p \left( \vect{y} \middle| \vect{X}^{(k)}, \thetamap \right)$ \tcp*[r]{Equation \eqref{Eq:model_C}}  
	}
	Construct sample-based approximation  $\tilde{p} \left( \vect{y} \middle| \bx, \mathcal{D} \right)$ using samples $\vect{y}^{(k)}, k=1,...,K$
\caption{Making predictions for new $\bx$ using the generative model }
\label{Alg:GeneralAlgPredictions}
\end{algorithm}

\subsubsection{Predictive performance metrics}
\label{sec:predmetrics}

For the purpose of quantifying the predictive performance, we consider a validation dataset $\mathcal{D}_{v} = \big\lbrace \vect{x}^{(i_v)}, \vect{y}^{(i_v)} \big\rbrace_{i_v=1}^{N_v}$ consisting of $N_{v}$ input-output pairs of the FGM, for which we evaluate the following two metrics using the posterior predictive density:

\begin{description}
\item[Coefficient of determination $R^2$] The coefficient of determination $R^2$ is a standard metric \citep{zhang2017coefficient}  which assesses the accuracy of point estimates, 
and in particular of the mean  $\bs{\mu}(\bx^{(i_v)})$ of the predictive posterior  of our trained model for each validation input $\vect{x}^{(i_v)}$ i.e.: 

\be
\bs{\mu}(\bx^{(i_v)})=\mathbb{E}_{p \left( \vect{y} \middle| \vect{x}^{(i_v)}, \mathcal{D} \right)} \left[ \vect{y} \right], \qquad i_v=1,\ldots,N_{v}.
\label{Eq:PosteriorPredMean}
\ee

The mean of the posterior predictive is estimated using Monte Carlo (see Algorithm \ref{Alg:GeneralAlgPredictions}) and is compared to the reference FGM outputs $\{ \vect{y}^{(i_v)} \}_{i=1}^{N_{\text{val} } }$ as follows:
\begin{align}
R^2 = 1 - \cfrac{ \sum_{i_v=1}^{N_v}
\left| \left| \vect{y}^{(i_v)} - \bs{\mu}(\bx^{(i_v)}) \right| \right|_2^2 
}{ \sum_{i_v=1}^{N_v}
\left| \left| \by^{(i_v)} - \by_v \right| \right|_2^2
}
\label{Eq:DefinitionR2}
\end{align}
where  $\by_v = \frac{1}{N_v} \sum_{i_v=1}^{N_v} \vect{y}^{(i_v)}$ is the sample average  of the validation dataset. It can be noted that $R^2$ attains its maximum value, i.e. $R^2 = 1$, when the mean predictive estimates coincide with the actual FGM outputs in the validation dataset and deviations from these are weighted by the variability of the validation data appearing in the denominator of \refeqp{Eq:DefinitionR2}.

\item[Logscore \textbf{$\LS$}] This metric assess not just point estimates of the predictive posterior but also the associated predictive uncertainty.
In particular and for the purpose of computing $\LS$ we approximate  the otherwise intractable $p \left( \by \middle| \bx^{(i_v)}, \mathcal{D} \right)$ in \refeqp{Eq:PostPred_approx} at each validation input $\bx^{(i_v)}$, by a  Gaussian with a mean equal to the actual mean of the predictive posterior  $\bs{\mu}(\bx^{(i_v)})$ (\refeqp{Eq:PosteriorPredMean} - estimated by Monte Carlo) and a diagonal covariance matrix $\bs{S}(\bx^{(i_v)})$ containing the actual variances (also estimated by Monte Carlo - see Algorithm \ref{Alg:GeneralAlgPredictions}), i.e.:
\be
\bs{S}(\bx^{(i_v)})= \text{diag} \left( \sigma^2_j \left( \bx^{(i_v)} \right) \right) , \qquad i_v=1,\ldots,N_{v}
\ee
where 
\be
 \sigma_j^2(\bx^{(i_v)})=\mathbb{E}_{p \left( \vect{y} \middle| \vect{x}^{(i_v)}, \mathcal{D} \right)} \left[ (y_j-\mu_j(\bx^{(i_v)}))^2 \right], \qquad i_v=1,\ldots,N_{v}.
\ee

Subsequently, $LS$ is evaluated as follows:
\be
LS= \frac{1}{N_v} \sum_{i_v=1}^{N_v} \log \mathcal{N}\left(\by^{(i_v)} ~| ~ \bs{\mu}(\bx^{(i_v)}), \bs{S}(\bx^{(i_v)}) \right).
\ee
One notes that high $LS$ values are achieved not only when the predictive mean $\bs{\mu}(\bx^{(i_t)})$ is close to the true $\by^{(i_t)}$ but also when the predictive uncertainty (as measured by the variances $\sigma_j^2(\bx^{(i_t)})$) is simultaneously as small as possible.  It can finally be shown \citep{grigo_bayesian_2019} that $LS$ approximates the Kullback-Leibler divergence between the true $\pi(\by|\bx)$ and the (Gaussian approximation of the) predictive posterior $p(\by|\bx, \mathcal{D})$ averaged over the true distribution, say $\pi(\bx)$, of the inputs. 
 
\end{description}


\section{Numerical Illustrations}
\label{sec:num}
We demonstrate the capabilities of the proposed framework in  discovering predictive,  probabilistic surrogates on a two-dimensional diffusion problem. 
In the sequel, we specify particular elements of the proposed model that were presented generically in the previous sections and concretize parametrizations and their meaning. The goals of the numerical illustrations are:

\bi 

\item to examine the effect of the number  labeled data $N_l$ which  are the most expensive to obtain and to assess whether the model can perform well under small $N_l$ (i.e. a few tens of FGM runs, section \ref{sec:effectdl} ). 

\item to assess the ability of the model to learn effective and interpretable CGMs that provide insight to the relevant features of the high-dimensional input $\bx$ which are predictive of the output $\by$ (section \ref{sec:effectdl}). 

\item to examine the effect of the {\em amount} of virtual observables $\mathcal{D}_{\vo}$ and assess  whether the model's predictive performance can be improved by increasing the number $N_{\vo}$ of such data (section \ref{subsec:study_virtual_obs}).

\item  to examine the effect of the {\em type} of virtual observables provided for training. In particular, we consider three different types (namely coarse-grained residuals, hybrid and potential energy) and assess the model's predictive performance for each one of those (section \ref{subsec:study_virtual_obs}). 

\item to examine the effect of unlabeled data $\mathcal{D}_u$ which are inexpensive to obtain and assess whether the model's predictive performance can be improved by increasing the number $N_u$ of such data (section \ref{subsec:supp_unlabeled_data}).

\item to examine the effect of the information bottleneck implied by the latent variables $\bz$ and the CGM  and to assess the effect of the the dimension of  $\vect{z}$ and the CGM's state variables (i.e. $\bxx$ and $\byy$) on the predictive performance of the model (section  \ref{sec:effectxx})

\item to assess the predictive performance of the model  under high-dimensional parametric inputs $\bx$  and under ``interpolative'' and ``extrapolative" conditions. The latter distinction refers to the ability to predict the (equally high-dimensional) output vector $\by$ under   boundary conditions that were used during training (interpolative) or not (extrapolative) (section \ref{sec:effectbc}). 

\item to investigate the efficiency and accuracy of the trained surrogate in a many-query application involving uncertainty propagation  (section \ref{subsection:uq}). 

\ei

Some of the simulation results  as well as the corresponding  code will be made available at the following github repository\footnote{\url{https://github.com/bdevl/PGMCPC}} upon publication.

\clearpage
\subsection{Definition of physical problem}

For the numerical illustration of our modeling framework we consider a linear elliptic PDE defined on the unit square $\Omega = \left[0,1 \right]^d$ in dimension $d=2$. We can write the governing equations as a two-field problem

\be
\textrm{conservation law:  }\nabla \cdot \bs{J}(\bs{s}) = f , \quad \forall \bs{s} \in \Omega \hfill 
\label{Eq:Poisson0} 
\ee 
\be
\textrm{constitutive law:  }\bs{J}(\bs{s}) = - \nabla \left( \kappa \left( \bs{s}\right) u(\bs{s}) \right)
 \quad \forall \bs{s} \in \Omega
 \label{Eq:Poisson1a} 
 \ee
 with boundary conditions:
 \be
u = u_D, \quad \vect{s} \in \Gamma_D
\label{Eq:Poisson2} 
\ee
\be
 \flux \cdot \vect{n} = \bs{0},  \quad  \vect{s} \in \Gamma_N
 \label{Eq:Poisson3}
\ee
where $u(\bs{s})$ is a scalar field to which one might attribute the physical meaning of temperature or pressure or concentration,  $\bs{J}(\bs{s})$ is a  vector field representing \textit{flux}, and $\vect{n}$ is the unit outward normal vector. $\Gamma_N$ denotes  the part of the boundary where  Neumann boundary conditions are prescribed  and is comprised of the top and bottom sides of the unit square $\Omega$ i.e. for $\{\bs{s}|s_2=0 \textrm{ or } s_2=1\}$.
At the remaining boundary  $\Gamma_D$, i.e. the left and right side of the domain, we introduce randomized boundary conditions of the form

\begin{align}
\label{Eq:randomized_dirichlet_bc}
\begin{split}
u_D \left( \vect{s} \right) &= a_0 \cdot s_2 + a_1 \left( 1 - s_2 \right)  \qquad \vect{s} \in \left\lbrace \vect{s} \middle| s_1 = 0 \right\rbrace
\\
 u_D \left( \vect{s} \right) &= a_2 \cdot s_2 + a_3 \left( 1 - s_2 \right) \qquad \vect{s} \in \left\lbrace \vect{s} \middle| s_1 = 1 \right\rbrace 
\end{split}
\end{align}

with $a_i \sim \mathcal{U} \left[ -0.5, 0.5 \right] $.  \\
We model $\kappa \left( \bs{s}\right)$ 
 with a log-normally distributed  random field i.e. $\kappa(\bs{s})=e^{\lambda(\bs{s})}$  where the underlying Gaussian field has a spatially constant mean $\mu_{\lambda}$  
  and a covariance $\mathcal{C}_{\lambda} \left( \vect{s}, \vect{s}' \right)$ function given by

\begin{align}
\mathcal{C}_{\lambda} \left( \vect{s}, \vect{s}' \right) = \sigma_{\lambda}^2 \cdot \exp \left(  - \frac{1}{2} \frac{\left| \left| \vect{s} - \vect{s}' \right| \right|_2^2}{ l_{\lambda}^2} \right).
\label{eq:covar}
\end{align} 

The following values were used for the parameters:  $\mu_{\lambda} = 0.4$, $\sigma_{\lambda} = 0.8$ and  $l_{\lambda} = 0.04$ or $0.15$ (depending on the resolution of the FGM). 
The resulting random field $\kappa(\bs{s})$ exhibits significant variability with a coefficient of variation of $~0.95$ and the small correlation lengths necessitate fine discretizations  resulting in a high-dimensional random input $\bx$. \\

The numerical solution of the governing equations is obtained using a standard  Finite Element (FE) schemes. For the purposes of our illustrations we consider the following two FE discretizations giving rise  to the fine-grained (FGM) and coarse-grained (CGM) models in the previous discussion:

\bi
\item[\textbf{FGM}] This employs  a fine(r) discretization using a regular grid of size $d_f \times d_f$\footnote{
The use of regular grids is pursued in order to enable the use of convolutional neural networks (CNNs) \citep{lecun1999object, krizhevsky2012imagenet} for the parameterized densities, enabling a parsimonious description of a complex hierarchy of features. We note that expressing physically meaningful spatio-(temporal) features on possibly non-regular and unstructured domains is a challenge in itself, but not the subject of this investigation. As such we have chosen to constrain ourselves to the representation of the random field on a regular grid, which enables the use of methods that have reached maturity due to their extensive use in computer vision.
}.
Our simulations  are based on $d_f=32$ (for  $l_{\lambda} = 0.15$) and $d_f=64$ (for $l_{\lambda} = 0.04$) giving rise to $dim(\bx)=dim(\by)=1024$ and $4096$ respectively.
The random field $\kappa(\bs{s})$ is discretized using piece-wise constant functions over each pixel and the vector $\bx$ represents the value of  $\kappa(\bs{s})$ at the centroid of each pixel. Hence  $dim(\bx)=d_f^2$.

In anticipation of the {\em virtual observables} that will be enforced and are discussed in more detail in section \ref{sec:Example_VO}, we review here the weak form of the governing PDE which, in view of Equation \eqref{Eq:Poisson0} and the boundary conditions in Equation \eqref{Eq:Poisson2} and Equation \eqref{Eq:Poisson3} becomes:
\be
 -\int_{\Omega} \nabla_s w \cdot \bs{J} ~d\bs{s} - \int_{\Omega} w ~f~d\bs{s} =0 
 \label{eq:wr1}
 \ee
 or upon making use of the constitutive equation (\ref{Eq:Poisson1a})
 \be
\int_{\Omega} \nabla_s w \cdot \kappa ~\nabla_s u~d\bs{s} - \int_{\Omega} w ~f~d\bs{s} =0.
\label{eq:wr2}
\ee
The  {\em admissible} weight functions   $w \in \mathcal{W}$  belong in the set $\mathcal{W}=\{ w(\bs{s})~ | ~w(\bs{s}) \in H^1(\Omega), w(\bs{s})=0 \textrm{ on }  \Gamma_D\}$. We denote by $\by$ the discretized representation of $u(\bs{s})$ with the usual FE  shape functions which, upon substitution in Equation \eqref{eq:wr2}, and for each $w \in \mathcal{W}$ yields a residual $r_{w}: \RR^{d_x} \times \RR^{d_y} \to \RR$:
\be
r_w(\by ; \bx)=0
\label{eq:resw}
\ee
We note that depending on the choice of the weight functions $w$ (at least) six methods (i.e. collocation, sub-domain, least-squares, (Petrov)-Galerkin, moments) arise as  special cases \citep{finlayson_method_1972}.

It is also well-known that the solution to this problem, as with many problems in computational physics,  can be obtained by minimizing an appropriate functional which in this case reduces to the potential energy function $\mathcal{V}$ given by:
\be
\mathcal{V}=\frac{1}{2} \int_{\Omega} \kappa \left| \nabla_s u \right|^2~d\bs{s} - \int_{\Omega} f~u~d\bs{s} 
\label{eq:potencont}
\ee
Upon discretization, this suggests that the  solution vector $\by$ can be found by minimizing  $V$ i.e.:
\be
 min_{\by} ~V(\by;\bx)
 \label{eq:poten}
 \ee
 where $V$ is the discretized potential energy obtained by using the discretized versions of $\kappa$ and $u$ in  $\mathcal{V}$ of Equation \eqref{eq:potencont}.

\begin{figure}[t!]
    \begin{center}
    \scalebox{1.1}{
\input{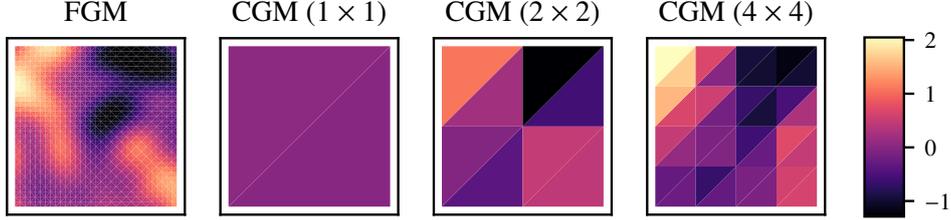}
}
    \caption{ Comparison of  a sample $\vect{x^{(i)}}$ of the discretized of the Gaussian random field $\lambda(\bs{s})$  of the FGM (left - \refeqp{eq:covar} with $l_{\lambda} = 0.15$) with the (log of the posterior mean of the) corresponding $\bxx^{(i)}$ for three different CGM discretizations i.e. $1\times 1$ , $2\times 2$ and $4\times 4$ (The posterior means $\mathbb{E} \left[ q \left( \vect{X}^{(i)} \right) \right]$ are based on $N_l = 512$ training data).
    The CGMs encode \textit{effective} properties $\vect{X}^{(i)}$ via the trained model density $p \left( \vect{X} \middle| \vect{x} \right)$. As the CGM is refined, it captures more details of the underlying FGM properties, e.g. areas in the problem domain with higher/lower conductivity $\bx$ in the FGM correspond to higher/lower values of  $\bxx$  in the CGM.
    }
    \label{fig:coarse_graining_illustration}
    \end{center}
\end{figure}

We note that the  output vector $\by$ which  corresponds to the discretization of $u(\bs{s})$  is of similar dimension $d_y = dim(\by)= ( d_f + 1 ) ^2$ as well\footnote{excluding boundary conditions} (Figure \ref{fig:coarse_graining_illustration}). We do not consider the discretization error of the FGM,  as our goal in this work is to predict $\by$ (i.e. the discretized solution), and as such assume it to be of sufficient accuracy.

\item[\textbf{CGM}] This is based  on a FE solver on a coarse(r) regular grid of size $d_c \times d_c$. Analogously to the FGM, the CGM input vector $\bxx$ represents the property within each of the  pixels and  is therefore of dimension $dim(\bxx)=d_c^2$. The FE solver yields the output vector $\byy$ (which  represents  $u(\bs{s})$) and is therefore of dimension $dim(\byy)=(d_c+1)^2$ as well  \footnote{excluding boundary conditions}.
Various values of $d_c$ were considered (see Figure \ref{fig:coarse_graining_illustration}) - in all cases $d_c << d_f$) in order to assess the effect of the dimensionality of the CGM in the predictive estimates.
We note that this particular form of the CGM was adopted for simplicity and due to the fact that boundary conditions can be readily incorporated in it rather than having to learn their effect as well (e.g. by including them in $\bx,\bxx$). Nevertheless, any coarse-grained or reduced-order model from the vast literature on this topic can be employed instead. 

\ei

\begin{figure}[p]
\begin{center}
\includegraphics[scale=0.75]{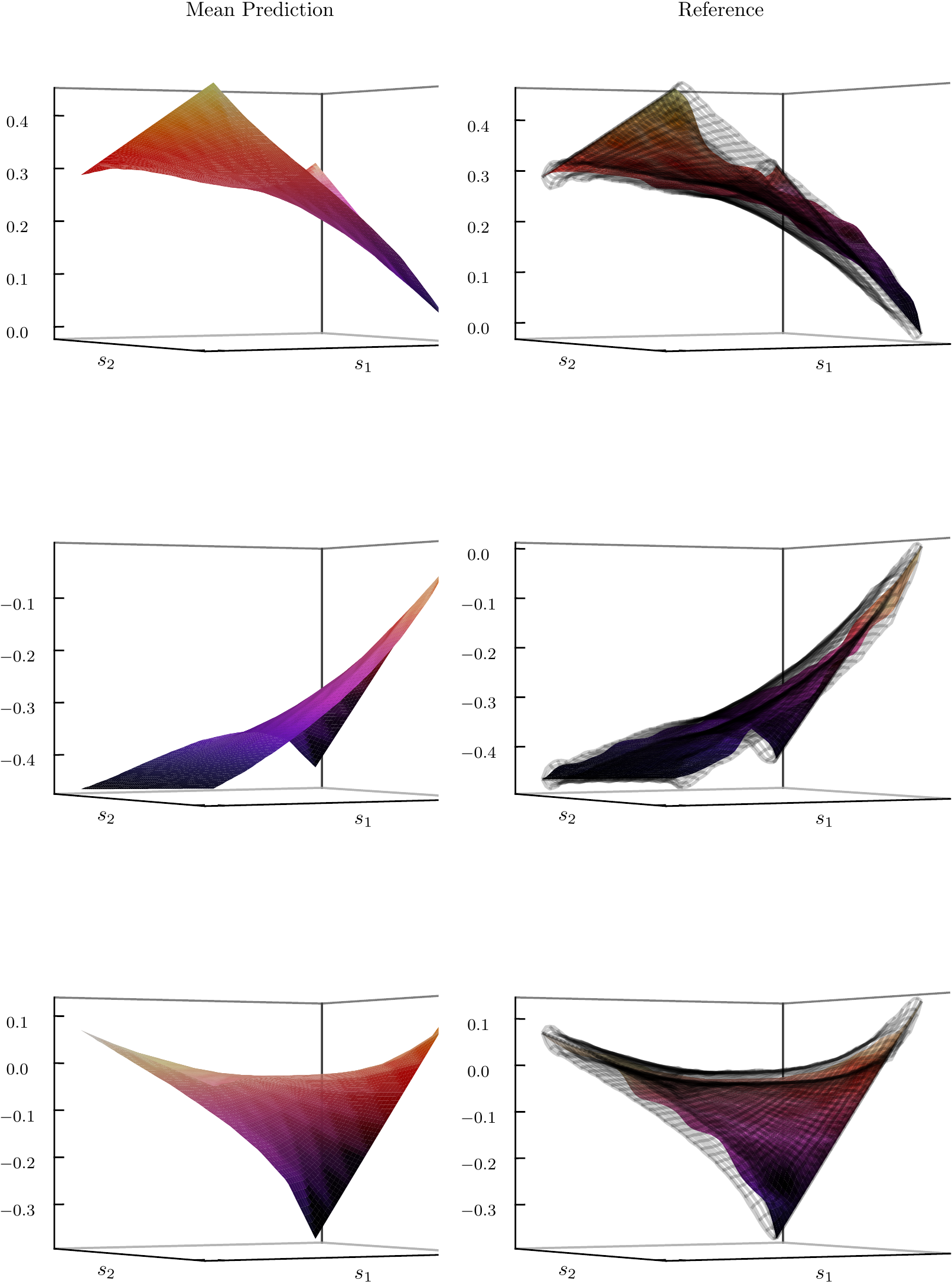}
\caption{The \textit{left} column provides examples of the mean of the posterior predictive $p \left( \vect{y} \middle| \vect{x}, \mathcal{D} \right)$ for various examples $\vect{x}$ not seen during training. On the \textit{right} we contrast this CGM based predictions with the actual solution obtained by solving the FGM (ground truth / reference), where the enveloping black mesh corresponds to the $90\%$ confidence interval of the posterior predictive. ($\left( 64  \times 64 \right)$ FGM, $\left( 8  \times 8  \right)$ CGM, $l_{\lambda} = 0.04$)}
\label{fig:predictions}
\end{center}
\end{figure}

\subsection{Specification of the generative model}
\label{sec:spec_gen_model}

Given the physical problem above and the definitions of the associated input $\bxx,\bx$ and output vectors $\byy,\by$, we provide details on the parameterization of the generative model which was generically described in section \ref{sec:method}. In particular, the following modeling choices were made:

\begin{enumerate}[(a)]

\item we employ a densely connected convolutional neural network \citep{huang2017densely} to parameterize the mean $\bs{f}(\bz; \bt_x )$ as well as the input-dependent diagonal covariance matrix $\mat{S}_{\vect{x}}(\bz; \bt_x )$  in Equation \eqref{Eq:model_A}. In addition, we make use of the same  architecture for the amortized encoder $q_{\Phi} \left( \vect{z} \middle| \vect{x} \right)$ (section \ref{sec:inference}). More specifically, the implementation is based on a variation  of the architecture proposed in \citep{zhu2018bayesian}. 
The  alterations  refer predominantly to a reduction in the complexity and expressivity since the latent space $\vect{z}$  encodes the salient features of  $\bx$, i.e. we only wish to retain information to the extent that it can help us in predicting effective properties by means of $p_{\vect{\theta}} \left( \vect{X} \middle| \vect{z} \right)$ (Equation \eqref{Eq:model_GP}).

\item The conditional density $\smash{\mathcal{N} \big( \vect{X} \big| \bs{g} \big(  \vect{z} ; \vect{\theta}_g \big), \mat{S}_{\vect{X}} \big)}$  defined by \refeqp{Eq:model_GP} relates the latent encoding $\vect{z}$ to the input $\bxx$ of the CGM (i.e. the apparent/effective/homogenized properties). The mean vector $\smash{\bs{g} \big(  \vect{z} ; \vect{\theta}_g \big)}$ depends on the latent variables $\vect{z}$ and is parameterized using a linear layer, i.e. $\smash{\bs{g} \big(  \vect{z} ; \vect{\theta}_g \big) = \mat{W}_g \vect{z} + \vect{b}_g}$ such that $\smash{\vect{\theta}_g = \big\lbrace \mat{W}_g, \vect{b}_g \big\rbrace}$, which was found to be most robust in the low-data regime (this could be trivially expanded to a shallow feedforward neural network).

\item For the dimension of the latent space we adopt the choice $ \text{dim} \left( \vect{z} \right) = 0.5 \cdot \text{dim} \left( \vect{X} \right)$. To motivate this choice, we note that the primary function of $\bz$ is to induce an information bottleneck which is able to retain information about \textit{effective} properties $\vect{X}$. A suitable choice however will always be problem-dependent (see also section \ref{sec:effectxx}).

\end{enumerate}

The general implementation of the model leverages and intertwines both Fenics \citep{logg2012automated} as well as PyTorch \citep{paszke2017automatic}. The CGM and its adjoint have been fully embedded within the automatic differentiation framework of PyTorch, enabling the fast and parallel solution of the CGM on the GPU (i.e. in batches).


\subsection{Virtual Observables} 
\label{sec:Example_VO}

Following the general discussion in section (\ref{sec:VirtualObservables}) on how domain knowledge can be introduced consistently in a probabilistic graphical model as artificial nodes (virtual observables), we discuss several types of such  virtual observables $\smash{\mathcal{D}_\vo}$ derived from the governing equations. 
We are primarily interested in those that can   inexpensively augment the training data and improve the predictive ability of the trained model even though they might provide \textit{incomplete} or \textit{partial} pieces of information at each input query point $\vect{x}^{(i_\vo)}$ about the underlying governing equations.
This property (partial information) will be reflected in the fact that most constraints we consider only carry information about a small subset of dimensions in the $\by$-space.
We note  that when  the virtual observables $\vect{o} \left( \vect{y}; \vect{x} \right)$  are linear with respect to $\by$, then  low-rank, closed-form updates for $\smash{   \big\lbrace q \big( \by^{(i_{\vo})}  \big) \big\rbrace_{i_{\vo}=1}^{N_{\vo}}  }$ (\refeqp{Eq:VariationalFactorization}) can be employed. Detailed information on these technical matters is provided in \ref{sec:virtual_obs_closed_form} and in the appendices referenced  in the ensuing discussion.

\begin{description}
\item[Weighted Residuals] As discussed in the previous section, the method of weighted residuals can be used to enforce the governing equations. Hence we propose using Equation \eqref{eq:wr2} as constraints that are probabilistically incorporated in the proposed model as discussed in section \ref{sec:VirtualObservables}. We note that the use of weighted residuals of PDEs has  also been advocated in deterministic machine-learning loss functions \citep{khodayi-mehr_varnet:_2019}. We  consider two categories of residuals $r_w(\by;\bx)$ based on two different types of weight functions $w$. The latter can be thought of as the lens through which the governing equations are viewed.

The first type, which we call \textbf{Coarse-Grained Residuals}, employs weight functions $w$ that correspond to the coarser discretization of the CGM. Due to the lower resolution of the corresponding mesh, they can be thought as enforcing the governing equations in a spatially-averaged sense. In particular and if we denote by $\bs{\Psi}(\bs{s})=\{\Psi_m(\bs{s})\}_{m_1=1}^{M_1}$ the vector containing the shape-function of the CGM, we consider $M_1$ weight functions $\{w_{m_1}\}_{m_1=1}^{M_1}$ of the form\footnote{We always ensure these are {\em admissible}.}:
\be
w_{m_1}(\bs{s})= \Psi_{m_1}(\bs{s}),  \quad m_1=1,\ldots,M_1
\label{eq:cgrwf}
\ee

The second type of residuals considered and which we call \textbf{Randomized Residuals} are based on using $M_2$ radial basis-type functions as weight functions $w$ i.e.

\be
w_{m_2}(\bs{s})= \exp \left( -\frac{ ||\bs{s}-\bs{s}_{0,m_2}||^2}{ \ell_{m_2}^2}  \right), \qquad m_2=1,\ldots,M_2
\label{eq:rbfwf}
\ee

The scale parameters $\{\ell_{m_2}\}_{m_2=1}^{M_2}$  were set equal to $0.1$ in subsequent investigations,  and the centers  $\smash{\{\bs{s}_{0,m_2}\}_{m_2=1}^{M_2}}$ are sampled uniformly over the problem domain i.e. $[0,1]^2$. 

 In contrast to the first type of residuals, these are capable of providing more localized information and over subdomains the size of which is determined by the scale parameters $\ell_{m_2}$ which can be adjusted accordingly. In the extreme where $\ell_{m_2} \to 0$, the weight function  $w_{m_2}$  becomes a Dirac-$\delta$ function and the corresponding constraint, a collocation-type one. The constraints associated with weighted residuals are enforced with infinite precision i.e. $\sigma_c =0$ in \refeqp{eq:volike}.

\begin{figure}[t!]
  \captionsetup{width=.85\linewidth}
    \begin{center}
    \scalebox{0.7}{
\input{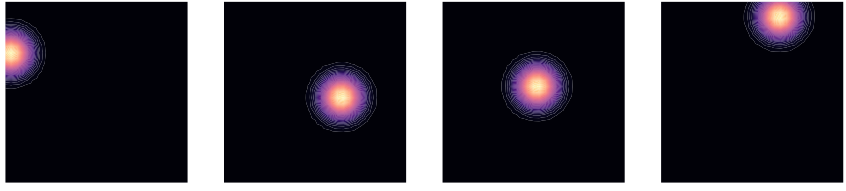}
}
    \caption{Illustration of $4$ randomly sampled radial basis-type weight functions (Eq. \eqref{eq:rbfwf}) corresponding to the  Randomized Residuals. Instead of randomly sampling collocation point at which the PDE is enforced, we randomly sample Galerkin weight functions that enforce  governing equations in a  spatially-averaged sense.}
    \label{Fig:IllustrationBasisFunctions}
    \end{center}
\end{figure}

\item[Conservation (Flux) Constraint] 
The second category  of constraints that we employ can also be cast as a special case of weighted residuals, but  operating instead directly on the conservation law (Equation \eqref{Eq:Poisson0}) i.e. on the flux variable $\bs{J}$ as in Equation \eqref{eq:wr1}.  In particular, we make use of indicator functions of subdomains $\Omega_{m_3} \subseteq \Omega$ as  weight functions $w_{m_3}$, i.e.:
\be
w_{m_3}(\bs{s})=1_{\Omega_{m_3}}(\bs{s}), \quad m_3=1,\ldots,M_3.
\label{eq:fluxwf}
\ee
We note that in this case, Equation \eqref{eq:wr1} reduces to:
\be
\int_{\pa \Omega_{m_3}} \bs{J} ~d\Gamma - \int_{\Omega_{m_3}} f~d\bs{s}=0
\label{eq:wr3}
\ee
where the first integration is over the boundary of $\Omega_{m_3}$.
The subdomains $\Omega_{m_3}$ are selected to coincide with the finite elements of the CGM (Figure \ref{fig:coarse_graining_illustration}). The flux $\bs{J}$ is computed using the constitutive law in Equation \eqref{Eq:Poisson1a} from the discretized solution vector $\by$.
Even  though the spatial resolution of the weight functions is analogous to the ones in the Coarse-Grained Residuals above, the information the residuals of Equation \eqref{eq:wr3} provide is of a different physical nature. Since not even the FGM  satisfies such flux constraint perfectly, we learn the precision  $\sigma_c^{-2}$ (\refeqp{eq:volike})  with which these constraints are enforced by introducing a prior that promotes larger values (\ref{App:FluxConstraint}). This is analogous to the well-known Automatic Relevance Determination  (ARD, \citep{bishop_pattern_2006}) on the associated constraints.

\item[Energy] The final constraint that we make use of pertains to the type presented in Equation \eqref{eq:potential} (section \ref{sec:VirtualObservables}) where the actual potential energy (Equation \eqref{eq:poten}) is employed. In contrast to the other constraints discussed, this provides  \textit{complete} information at each input query point, i.e. by minimizing $V$ which implies fully enforcing the corresponding virtual observable, one can perfectly determine the solution vector $\by$. This precludes low-rank updates and makes the incorporation of this constraint  more expensive. We provide details on how  $\smash{\big\lbrace q \big( \by^{(i_{\vo})}  \big) \big\rbrace_{i_{\vo}=1}^{N_{\vo}}}$  is updated using stochastic second-order optimization in \ref{sec:PotFctVirtObs}.

 \end{description}


\subsection{Predictive performance and the effect of $N_l$}
\label{sec:effectdl}

In the simplest scenario, the model is given access solely to a set of labeled data $\smash{\mathcal{D}_l = \big\lbrace \vect{x}^{(i_l)}, \vect{y}^{(i_l)} \big\rbrace_{i_l=1}^{N_l} }$ (i.e. $N_u=N_{\vo}=0$). In the following we demonstrate as a baseline that the model generalizes well in the {\em Small labeled Data} regime, owing to the use of the information-bottleneck variables $\bz$ as well as the CGM.

As observed in Figure \ref{fig:labeled_data_saturation}, the model achieves very high scores with only $N_l =128$ labeled data in terms of the $R^2$ (the largest possible value of $R^2$ is $1$) and $N_l=64$ in terms of the $LS$ score.  We observe that  further increase of $N_l$ results in minimal if not negligible improvement i.e. the model has saturated. While alterations in the neural networks involved can be expected to change the particular values, we note that the saturation effect is a consequence of the limited capacity of the CGM which lies at the center of the model proposed. That is, for a given a CGM with the optimal values for its parameters, we can only predict the FGM output $\by$ up to a certain level of detail. Hence even if infinite (labeled) data were available, the predictive scores of the model would not improve further and the remaining pieces would be enveloped by the predictive uncertainty (see Figure \ref{fig:predictions}).
On the other hand, if the CGM was removed and was substituted  by a more expressive (and with more parameters) black-box model (e.g. another neural net), its predictive performance would not be as high with so few labeled data but would continue to increase (as much as its capacity would allow) with increasing $N_l$. This saturation effect arising from the CGM has also been observed in the discriminative model proposed in \citep{grigo_physics-aware_2019} where procedures for the adaptive refinement of the CGM 
were proposed. These were driven by the ELBO $\mathcal{F}$, which provides a natural score function for each model,  but were not pursued in this work.

\setlength{\figWidth}{0.65\textwidth}
\setlength{\figHeight}{0.32\textheight}

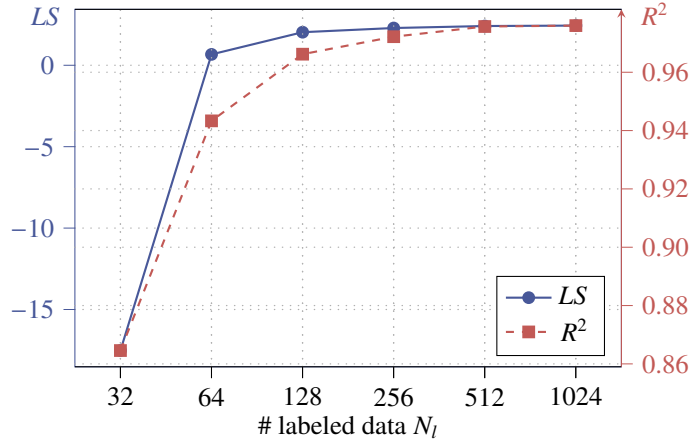
\begin{figure}[!htbp]
    \centering
\begin{tikzpicture}


\pgfplotsset{
  grid style = {
    dash pattern = on 0.05mm off 1mm,
    line cap = round,
    black,
    line width = 0.5pt
  }
}

\begin{axis}[
xmode = log,
log basis x={2},
grid=major,
height=\figHeight,
tick align=outside,
tick pos=left,
width=\figWidth,
x grid style={white!69.0196078431373!black},
xlabel={\# labeled data $N_l$},
xtick style={color=black},
y grid style={white!69.0196078431373!black},
ylabel={\textcolor{color0}{$LS$}},
ymin=-18.5185902450085, ymax=3.4345972058773,
ytick style={color=color0},
y tick label style={color=color0},
y axis line style={color=color0},
xtick={},
xticklabels={},
ymajorgrids,
xmajorgrids,
y label style={at={(axis description cs:+0.12,0.93)},rotate=-90,anchor=south},
]
\addplot [thick, color0, mark=*, mark size=2, mark options={solid}] 
table {
32 -17.52071808815
64 0.667488110661507
128 2.03075416088104
256 2.28788646459579
512 2.40975155115128
1024 2.43672504901886
}; 

\end{axis}

\begin{axis}[
xmode = log,
log basis x={2},
axis y line=right,
height=\figHeight,
tick align=outside,
title={ },
width=\figWidth,
x grid style={white!69.0196078431373!black},
xtick pos=left,
xtick style={color=black},
y grid style={white!69.0196078431373!black},
ylabel={\textcolor{color1}{$R^2$}},
ymin=0.859004994869232, ymax=0.98156765794754,
ytick pos=right,
ytick style={color=black},
ytick={0.84,0.86,0.88,0.9,0.92,0.94,0.96},
yticklabels={0.84,0.86,0.88,0.90,0.92,0.94,0.96},
xtick={32, 64, 128, 256, 512, 1024},
xticklabels={32, 64, 128, 256, 512, 1024},
ytick style={color=color1},
y tick label style={color=color1},
y axis line style={color=color1},
ymajorgrids,
xmajorgrids,
legend pos=south east,
y label style={at={(axis description cs:+0.9,0.93)},rotate=-90,anchor=south},
]
\addlegendimage{/pgfplots/refstyle=LogscorePlot}\addlegendentry{$LS$}
\addplot [dashed,thick, color1, mark=square*, mark size=2, mark options={solid}]
table {%
32 0.864576025009155
64 0.943331642150879
128 0.9661925137043
256 0.972190955877304
512 0.975679172277451
1024 0.975996627807617
}; 
\addlegendentry{$R^2$}

\end{axis}

\end{tikzpicture}
    \caption{Predictive performance in terms of the $R^2$ and $LS$ metrics   as a function of the number of labeled data points $N_l$ ($N_u=N_{\vo}=0$).  \resamplingstatement}
    \label{fig:labeled_data_saturation}
\end{figure}

\FloatBarrier


\subsection{Effect of the amount and type of  {\em virtual observables}}
\label{subsec:study_virtual_obs}

In the following,  we  demonstrate the benefits of the inclusion of virtual observables to the predictive performance of the proposed model.
In order to quantify this benefit, we consider  the posterior predictive density $p \left( \vect{y} \middle| \vect{x} , \mathcal{D}_\labeled, \mathcal{D}_\vo \right)$  (section \ref{sec:predictions}) 
as a function of labeled data $\mathcal{D}_l$ as well as of the  virtual observables $\mathcal{D}_{\vo} = \smash{ \big\{ \vect{x}^{(i)}, \obs^{(i)} ) \big\}_{i=1}^{N_{\vo}}}$. 
We omit in these experiments, {\em unlabeled data} $\mathcal{D}_u$ (i.e. $N_u=0$), the effect of which will be examined in section \ref{subsec:supp_unlabeled_data}.
In particular, we examine the improvement in the  predictive performance,  i.e. in the metrics $R^2$ and $LS$ (section \ref{sec:predmetrics}),  of the three baseline models (for $N_{\vo}=0$) corresponding to the following number of  labeled data i.e.:
\be
N_l= \left\lbrace 16, ~ 32, ~ 64 \right\rbrace
\ee
when  $N_{\vo}$ virtual observables are added, where: 
\be
N_{\vo} = \left\lbrace  32, ~ 64, ~ 128, ~ 196, ~ 256 \right\rbrace
\ee
Furthermore, we examine the effect of the different types of virtual observables by  considering the following three categories:
\bi
\item \textbf{CGR}: At each input query point $\vect{x}^{(i_\mathcal{O})}$, $M_1=25$ Coarse-Grained Residuals (\refeqp{eq:cgrwf}) are  observed .
\item \textbf{Hybrid}: At each input query point   $\vect{x}^{(i_\mathcal{O})}$ the CGR ($M_1 = 25$), a set of Randomized weighted residuals ($M_2=60$, \refeqp{eq:rbfwf}) and the conservation of flux ($M_3=32$, Equation \eqref{eq:fluxwf}) are observed. 
\item \textbf{Energy}: At each input  query point $\vect{x}^{(i_\mathcal{O})}$  the potential energy is observed.
\ei

we report on results in Figure Figure \ref{fig:vo_combined_visualizations}, where the left column depicts the evolution of the $R^2$ and $LS$ for different values of $N_{\vo}$ and for virtual observables of the CGR type.
One can readily observe that, for all three $N_l$  values (i.e. number of labeled data), the introduction of the domain-knowledge  in the form of these residual-type constraints leads to a significant improvement of the model's predictive accuracy. 
Furthermore, with the virtual observables introduced, one can attain with only $N_l=16$,  scores that in Figure \ref{fig:labeled_data_saturation} required $N_l=512$ labeled data i.e. a significant reduction in the number of times the FGM needs to be solved.
As one would perhaps expect, the gains from the virtual observables are  more pronounced for small numbers of labeled data i.e. when the model still struggles to generalize based on the too few labeled data points and therefore has more room to improve.
Despite the fact that these virtual observations  $\vect{\hat{o}} \in \mathbb{R}^{32}$ only provide partial information, the model is still able to leverage this to improve upon its predictive performance.

In the right column of Figure \ref{fig:vo_combined_visualizations} we expand upon these results by  considering \textit{different types} of virtual observables and by quantifying the impact of their informational content on the model's predictive performance.
We note that the energy virtual observables have the most striking benefit which was to be expected as they provide  complete information on the associated FGM output. 
Secondly, the {\em Hybrid}-type seems to yield  a higher improvement in the model's predictive score as compared to the CGM-type.
Finally in Figure \ref{fig:vo_gridplot}, we provide additional details by depicting  the $LS$ metric as a function of both $N_{\vo}$ and $N_l$.


\setlength{\figWidth}{0.54\textwidth}
\setlength{\figHeight}{0.31\textheight}

    \begin{figure}[!ht]
\centering
    \begin{subfigure}{0.45\linewidth}
        \centering
        \caption{$N_l = 16 ~~|$  CGR}
                \vspace{-0.15cm}
\begin{tikzpicture}[scale=0.8,trim axis right,trim axis left]

\pgfplotsset{
  grid style = {
    dash pattern = on 0.05mm off 1mm,
    line cap = round,
    black,
    line width = 0.85pt
  }
}

\begin{axis}[
xtick = {},
xticklabels = {},
grid=major,
height=\figHeight,
tick align=outside,
tick pos=left,
width=\figWidth,
x grid style={white!69.0196078431373!black},
xlabel={\# virtual observables $N_{\mathcal{O}}$},
xmin=-6, xmax=267.2,
xtick style={color=black},
y grid style={white!69.0196078431373!black},
y label style={at={(axis description cs:+0.15,0.975)},rotate=-90,anchor=south},
ylabel={\textcolor{color0}{$LS$}},
ymin=-185.53105584798236, ymax=20.79451091057899,
ytick style={color=color0},
y tick label style={color=color0},
y axis line style={color=color0},
ymajorgrids,
xmajorgrids,
]
\addplot [thick, color0, mark=*, mark size=2, mark options={solid}] 
table {
0 -176.158938789368
32 -8.69333036261869
64 -1.12082923600641
128 1.88826137528275
196 2.29428210980964
256 2.27716386101463
}; 

\end{axis}

\begin{axis}[
axis y line=right,
height=\figHeight,
tick align=outside,
title={ },
width=\figWidth,
x grid style={white!69.0196078431373!black},
xmin=-6, xmax=267.2,
xtick pos=left,
xtick style={color=black},
y grid style={white!69.0196078431373!black},
ylabel={\textcolor{color1}{$R^2$}},
ymin=0.78463030496201, ymax=0.985,
ytick pos=right,
ytick style={color=black},
y label style={at={(axis description cs:+0.865,0.975)},rotate=-90,anchor=south},
ytick style={color=color1},
y tick label style={color=color1},
y axis line style={color=color1},
ymajorgrids,
xmajorgrids,
legend pos=south east,
]
\addlegendimage{/pgfplots/refstyle=LogscorePlot}\addlegendentry{$LS$}
\addplot [dashed,thick, color1, mark=square*, mark size=2, mark options={solid}]
table {%
0 0.814217004179955
32 0.935035965887897
64 0.960747054183339
128 0.973832354039857
196 0.976368437752579
256 0.977664934143876
}; 
\addlegendentry{$R^2$}

\end{axis}

\end{tikzpicture}%
    \end{subfigure}
\hfil
    \begin{subfigure}{0.45\linewidth}
        \centering
    \addtocounter{subfigure}{+2}
        \caption{$N_l = 16 ~~|$  Comparison}
                \vspace{-0.15cm}
\begin{tikzpicture}[scale=0.8, trim axis right,trim axis left]

\pgfplotsset{
  grid style = {
    dash pattern = on 0.05mm off 1mm,
    line cap = round,
    black,
    line width = 0.85pt
  }
}

\begin{axis}[
height=\figHeight,
legend cell align={left},
legend style={fill opacity=0.8, draw opacity=1, text opacity=1, at={(0.97,0.03)}, anchor=south east, draw=white!80!black},
tick align=outside,
tick pos=left,
ytick={-10, -7.5, -5.0, -2.5, 0.0},
yticklabels={-10, -7.5, -5.0, -2.5, 0.0},
width=\figWidth,
x grid style={white!69.0196078431373!black},
xlabel={\# virtual observables $N_{\mathcal{O}}$},
xmajorgrids,
xmin=20.8, xmax=267.2,
xtick style={color=black},
y grid style={white!69.0196078431373!black},
ylabel={$\LS$},
ymajorgrids,
y label style={at={(axis description cs:+0.15,0.975)},rotate=-90,anchor=south},
ytick style={color=black}
]

\addplot [semithick, color2, mark=square*, mark size=2, mark options={solid}]
table {%
32 -1.16252580991676
64 1.43625138007825
128 2.1300278102198
196 2.29889474518887
256 2.32477278523631
};
\addlegendentry{Energy}

\addplot [semithick, color1, mark=triangle*, mark size=2, mark options={solid}]
table {%
32 -4.73006788377436
64 0.614160967678387
128 1.99937600534667
196 2.24854784254815
256 2.32829369750677
};
\addlegendentry{Hybrid}

\addplot [semithick, color0, mark=*, mark size=2, mark options={solid}]
table {%
32 -8.69333036261869
64 -1.12082923600641
128 1.88826137528275
196 2.29428210980964
256 2.27716386101463
};
\addlegendentry{CGR}

\end{axis}

\end{tikzpicture}
    \end{subfigure}
    
    \vspace{0.5cm}

    \begin{subfigure}{0.45\linewidth}
    \centering
    \addtocounter{subfigure}{-3}
        \caption{$N_l = 32 ~~|$  CGR}
                \vspace{-0.15cm}
\begin{tikzpicture}[scale=0.8, trim axis right,trim axis left]

\pgfplotsset{
  grid style = {
    dash pattern = on 0.05mm off 1mm,
    line cap = round,
    black,
    line width = 0.85pt
  }
}

\begin{axis}[
xtick = {},
xticklabels = {},
grid=major,
height=\figHeight,
tick align=outside,
tick pos=left,
width=\figWidth,
x grid style={white!69.0196078431373!black},
xlabel={\# virtual observables $N_{\mathcal{O}}$},
xmin=-6, xmax=267.2,
xtick style={color=black},
y grid style={white!69.0196078431373!black},
ylabel={\textcolor{color0}{$LS$}},
y label style={at={(axis description cs:+0.15,0.975)},rotate=-90,anchor=south},
ytick style={color=color0},
y tick label style={color=color0},
y axis line style={color=color0},
ymajorgrids,
xmajorgrids,
]
\addplot [thick, color0, mark=*, mark size=2, mark options={solid}] 
table {
0 -20.6432327826818
32 -0.14831018748157
64 0.968061449581927
128 1.63936767794869
196 1.9698685241468
256 2.24768038374005
}; 

\end{axis}

\begin{axis}[
axis y line=right,
height=\figHeight,
tick align=outside,
title={ },
width=\figWidth,
x grid style={white!69.0196078431373!black},
xmin=-6, xmax=267.2,
xtick pos=left,
xtick style={color=black},
y grid style={white!69.0196078431373!black},
ylabel={\textcolor{color1}{$R^2$}},
ymin=0.84, ymax=0.99,
ytick pos=right,
ytick style={color=black},
y label style={at={(axis description cs:+0.865,0.975)},rotate=-90,anchor=south},
ytick style={color=color1},
y tick label style={color=color1},
y axis line style={color=color1},
ymajorgrids,
xmajorgrids,
legend pos=south east,
]
\addlegendimage{/pgfplots/refstyle=LogscorePlot}\addlegendentry{$LS$}
\addplot [dashed,thick, color1, mark=square*, mark size=2, mark options={solid}]
table {%
0 0.879899722834428
32 0.954705253514377
64 0.96722719633218
128 0.97550032897429
196 0.976471799792665
256 0.97837567618399
}; 
\addlegendentry{$R^2$}

\end{axis}

\end{tikzpicture}%
    \end{subfigure}
\hfil
    \begin{subfigure}{0.45\linewidth}
    \centering
    \addtocounter{subfigure}{+2}
        \caption{$N_l = 32 ~~|$  Comparison}
                \vspace{-0.15cm}
\begin{tikzpicture}[scale=0.8, trim axis right,trim axis left]

\pgfplotsset{
  grid style = {
    dash pattern = on 0.05mm off 1mm,
    line cap = round,
    black,
    line width = 0.85pt
  }
}

\begin{axis}[
height=\figHeight,
legend cell align={left},
legend style={fill opacity=0.8, draw opacity=1, text opacity=1, at={(0.97,0.03)}, anchor=south east, draw=white!80!black},
tick align=outside,
tick pos=left,
ytick={0, 0.5, 1, 1.5, 2.0},
yticklabels={0, 0.5, 1, 1.5, 2.0},
width=\figWidth,
x grid style={white!69.0196078431373!black},
xlabel={\# virtual observables $N_{\mathcal{O}}$},
xmajorgrids,
xmin=20.8, xmax=267.2,
xtick style={color=black},
y grid style={white!69.0196078431373!black},
ylabel={$\LS$},
ymajorgrids,
y label style={at={(axis description cs:+0.15,0.975)},rotate=-90,anchor=south},
ytick style={color=black}
]

\addplot [semithick, color2, mark=square*, mark size=2, mark options={solid}]
table {%
32 0.694276209223655
64 1.70476233920743
128 2.19411307418501
196 2.32240685240015
256 2.36235710779826
};
\addlegendentry{Energy}

\addplot [semithick, color1, mark=triangle*, mark size=2, mark options={solid}]
table {%
32 0.0361889203682354
64 1.3389407856639
128 1.95297886556195
196 2.14856046931759
256 2.2979512609494
};
\addlegendentry{Hybrid}

\addplot [semithick, color0, mark=*, mark size=2, mark options={solid}]
table {%
32 -0.14831018748157
64 0.968061449581927
128 1.63936767794869
196 1.9698685241468
256 2.24768038374005
};
\addlegendentry{CGR}

\end{axis}

\end{tikzpicture}
    \end{subfigure}

        \vspace{0.5cm}

    \begin{subfigure}{0.45\linewidth}
    \centering
    \addtocounter{subfigure}{-3}
        \caption{$N_l = 64 ~~|$  CGR}
                \vspace{-0.15cm}
\begin{tikzpicture}[scale=0.8, trim axis right,trim axis left]

\pgfplotsset{
  grid style = {
    dash pattern = on 0.05mm off 1mm,
    line cap = round,
    black,
    line width = 0.85pt
  }
}

\begin{axis}[
xtick = {},
xticklabels = {},
grid=major,
height=\figHeight,
tick align=outside,
tick pos=left,
width=\figWidth,
x grid style={white!69.0196078431373!black},
xlabel={\# virtual observables $N_{\mathcal{O}}$},
xmin=-6, xmax=267.2,
xtick style={color=black},
y grid style={white!69.0196078431373!black},
ylabel={\textcolor{color0}{$\LS$}},
y label style={at={(axis description cs:+0.15,0.975)},rotate=-90,anchor=south},
ytick style={color=color0},
y tick label style={color=color0},
y axis line style={color=color0},
ymajorgrids,
xmajorgrids,
]
\addplot [thick, color0, mark=*, mark size=2, mark options={solid}] 
table {
0 0.861996034085751
32 1.46817992022543
64 1.80246275013143
128 1.94830402859881
196 1.96207744520415
256 2.10767440076144
}; 

\end{axis}

\begin{axis}[
axis y line=right,
height=\figHeight,
tick align=outside,
title={ },
width=\figWidth,
x grid style={white!69.0196078431373!black},
ytick={0.94, 0.95, 0.96, 0.97, 0.98},
yticklabels={0.94, 0.95, 0.96, 0.97, 0.98},
xmin=-6, xmax=267.2,
xtick pos=left,
xtick style={color=black},
y grid style={white!69.0196078431373!black},
ylabel={\textcolor{color1}{$R^2$}},
ymin=0.944129438271123, ymax=0.98521872862112,
ytick pos=right,
ytick style={color=black},
y label style={at={(axis description cs:+0.865,0.975)},rotate=-90,anchor=south},
ytick style={color=color1},
y tick label style={color=color1},
y axis line style={color=color1},
ymajorgrids,
xmajorgrids,
legend pos=south east,
]
\addlegendimage{/pgfplots/refstyle=LogscorePlot}\addlegendentry{$LS$}
\addplot [dashed,thick, color1, mark=square*, mark size=2, mark options={solid}]
table {%
0 0.953271842002869
32 0.965739950266751
64 0.972209992192008
128 0.97634538587617
196 0.976451864782369
256 0.979170650431195
}; 
\addlegendentry{$R^2$}

\end{axis}

\end{tikzpicture}%
    \end{subfigure}
\hfil
    \begin{subfigure}{0.45\linewidth}
    \centering
    \addtocounter{subfigure}{+2}
        \caption{$N_l = 64 ~~|$  Comparison}
        \vspace{-0.15cm}
\begin{tikzpicture}[scale=0.8, trim axis right,trim axis left]

\pgfplotsset{
  grid style = {
    dash pattern = on 0.05mm off 1mm,
    line cap = round,
    black,
    line width = 0.85pt
  }
}

\begin{axis}[
height=\figHeight,
legend cell align={left},
legend style={fill opacity=0.8, draw opacity=1, text opacity=1, at={(0.97,0.03)}, anchor=south east, draw=white!80!black},
tick align=outside,
tick pos=left,
width=\figWidth,
ytick={1.6, 1.8, 2.0, 2.2},
yticklabels={1.6, 1.8, 2.0, 2.2},
x grid style={white!69.0196078431373!black},
xlabel={\# virtual observables $N_{\mathcal{O}}$},
xmajorgrids,
xmin=20.8, xmax=267.2,
xtick style={color=black},
y grid style={white!69.0196078431373!black},
ylabel={$\LS$},
ymajorgrids,
y label style={at={(axis description cs:+0.15,0.975)},rotate=-90,anchor=south},
ytick style={color=black}
]

\addplot [semithick, color2, mark=square*, mark size=2, mark options={solid}]
table {%
32 1.69119970721583
64 2.00443605597942
128 2.25668866138954
196 2.33315439286985
256 2.35417145572297
};
\addlegendentry{Energy}

\addplot [semithick, color1, mark=triangle*, mark size=2, mark options={solid}]
table {%
32 1.53594680690462
64 1.83334678525378
128 2.08481238981721
196 2.2015640066516
256 2.23050471068987
};
\addlegendentry{Hybrid}

\addplot [semithick, color0, mark=*, mark size=2, mark options={solid}]
table {%
32 1.46817992022543
64 1.80246275013143
128 1.94830402859881
196 1.96207744520415
256 2.10767440076144
};
\addlegendentry{CGR}

\end{axis}

\end{tikzpicture}
    \end{subfigure}
    \vspace{0.5cm}
\caption{\textsc{Left Column:} Predictive performance of a model trained on $N_l$ labeled data, $N_\vo$ virtual observables of type CGR ($N_\unlabeled=0)$. \textsc{Right Column:} Comparison of predictive performance in terms of the $LS$ metric  with respect to 3  different \textit{types} of virtual observables. The baseline performance for $N_\vo = 0$ has been removed to improve clarity but the corresponding values can be found in Figure \ref{fig:labeled_data_saturation}.  \resamplingstatement}
    \label{fig:vo_combined_visualizations}
    \end{figure}
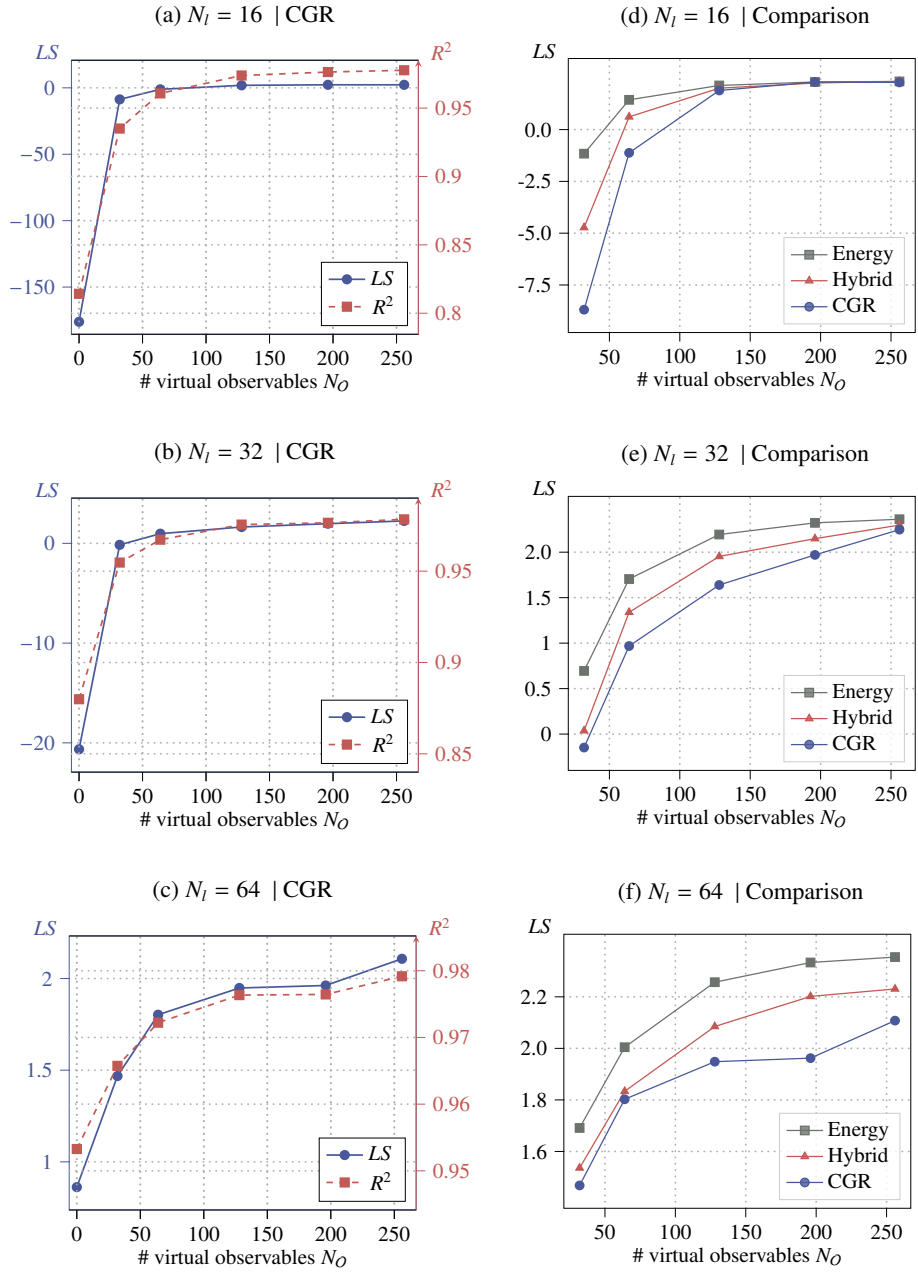



\begin{figure}
    \centering
    \input{gfx/virtualobservables/gridplot2/gridplot_t.pgf}
    \caption{$LS$ Score as function of $N_l$ (number of labeled data) and ($N_\vo = 0$ not shown to make results more clearly visible). \resamplingstatement}
    \label{fig:vo_gridplot}
\end{figure}


\clearpage
\subsection{Effect of unlabeled data}
\label{subsec:supp_unlabeled_data}

In this section, we study the effect of unlabeled data $\mathcal{D}_u= \big\lbrace \vect{x}^{(i)} \big\rbrace_{i=1}^{N_\unlabeled}$, i.e. semi-supervised learning, in the model's predictive accuracy.
To this end we investigate the predictive posterior $p \left( \vect{y} \middle| \vect{x}, \mathcal{D}_\unlabeled, \mathcal{D}_\labeled \right)$ as the number of unlabeled data $N_u$ increases. At the end of the section we consider simultaneously virtual observables $\mathcal{D}_{\vo}$ and assess their combined effect.
mainWe re-emphasize that unlabeled data are inexpensive to obtain (i.e. just inputs) and if the generative model proposed can exploit their informational context in improving its predictive ability, this would be of high utility.

In Figure \ref{Fig:UnsupervisedDataEffect} we present the evolution of predictive metrics $R^2$ and $LS$ as a function of the number of labeled data $N_l$ for two models. The blue line corresponds to no unlabeled data i.e. $N_u=0$ whereas the  red line to  $N_u=256$ such data.
In both Figures the benefit of $\mathcal{D}_{u}$ can be clearly observed.  The unlabeled data contribute  in  the identification of the lower-dimensional encoding $\bz$, i.e. a compressed description of the input $\bx$ which in turn informs the prediction of the output $\by$ through $\bxx$ i.e. the  CGM (Figure \ref{fig:pgmfull}). As one can also observe, the benefit of unlabeled data decreases the higher $N_l$ (i.e. the number of labeled data) is. This is not  unexpected as the room for improvement is smaller for higher $N_l$.

\setlength{\figHeight}{6.5cm}
\setlength{\figWidth}{8cm}

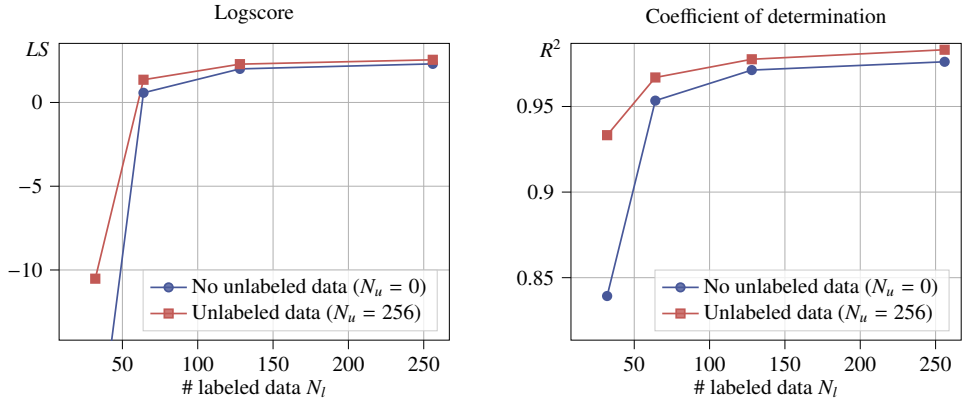
\begin{figure}[h!]
\begin{center}

\begin{tikzpicture}[scale=0.8]


\begin{groupplot}[group style={group size=2 by 1, horizontal sep=2cm}]
\nextgroupplot[
y label style={at={(axis description cs:+0.14,0.93)},rotate=-90,anchor=south},
height=\figHeight,
legend cell align={left},
legend style={fill opacity=0.8, draw opacity=1, text opacity=1, at={(0.97,0.03)}, anchor=south east, draw=white!80!black},
tick align=outside,
tick pos=left,
title={Logscore},
width=\figWidth,
x grid style={white!69.0196078431373!black},
xlabel={\# labeled data $N_l$},
xmajorgrids,
xmin=8, xmax=267,
xtick style={color=black},
y grid style={white!69.0196078431373!black},
ylabel={$\LS$},
ymajorgrids,
ymin=-14.1887160023053, ymax=3.52605744335386,
ytick style={color=black}
]
\addplot [thick, color0, mark=*, mark size=2, mark options={solid}]
table {%
32 -21.8085345219875
64 0.576718978640357
128 2.00294132637126
256 2.30298050887444
};
\addlegendentry{No unlabeled data ($N_u = 0$)}
\addplot [thick, color1, mark=square*, mark size=2, mark options={solid}]
table {%
32 -10.5197917808379
64 1.35599973583095
128 2.2868405619677
256 2.54695503888307
};
\addlegendentry{Unlabeled data ($N_u = 256$)}


\nextgroupplot[
y label style={at={(axis description cs:+0.14,0.93)},rotate=-90,anchor=south},
height=\figHeight,
legend cell align={left},
legend style={fill opacity=0.8, draw opacity=1, text opacity=1, at={(0.97,0.03)}, anchor=south east, draw=white!80!black},
tick align=outside,
tick pos=left,
title={Coefficient of determination},
width=\figWidth,
x grid style={white!69.0196078431373!black},
xlabel={\# labeled data $N_l$},
xmajorgrids,
xmin=8, xmax=267,
xtick style={color=black},
y grid style={white!69.0196078431373!black},
ylabel={$R^2$},
ymajorgrids,
ymin=0.813511683444182, ymax=0.98686731374926,
ytick style={color=black}
]
\addplot [thick, color0, mark=*, mark size=2, mark options={solid}]
table {%
32 0.83928838285847
64 0.953422106163842
128 0.971270945242473
256 0.976112308309359
};
\addlegendentry{No unlabeled data ($N_u = 0$)}
\addplot [thick, color1, mark=square*, mark size=2, mark options={solid}]
table {%
32 0.933239482130323
64 0.966948686753
128 0.977587643330985
256 0.983148347006904
};
\addlegendentry{Unlabeled data ($N_\unlabeled = 256$)}
\end{groupplot}

\end{tikzpicture}
\end{center}
\caption{A model trained on  a certain number of labeled data $N_\labeled$ is compared to a model which in addition had access to $N_\unlabeled=256$ unlabeled data points. The latter achieved consistently better performance. The weight afforded to unlabeled and unlabeled data in the objective (i.e. ELBO) has been normalized to be equal. \resamplingstatement}
\label{Fig:UnsupervisedDataEffect}
\end{figure}
 
Figure \ref{fig:effect_unlabeled_data} conveys similar information by varying the number of unlabeled data points while $N_l$ if fixed (either to $N_l=32$ or $N_l=64$). The improvement in the  predictive performance due to addition of unlabeled data points can be clearly observed. We further note that this improvement is always less than what one would attain with additional labeled data or with virtual observables (Figure \ref{fig:vo_gridplot}).

\setlength{\figHeight}{6.5cm}
\setlength{\figWidth}{8cm}

\begin{figure}
    \centering
    \begin{tikzpicture}[scale=0.8]


\begin{groupplot}[group style={group size=2 by 1, horizontal sep=2cm, vertical sep=2cm}]

\nextgroupplot[
height=\figHeight,
legend cell align={left},
legend style={fill opacity=0.8, draw opacity=1, text opacity=1, at={(0.97,0.03)}, anchor=south east, draw=white!80!black},
tick align=outside,
tick pos=left,
title={\# labeled data $N_l = 32$},
width=\figWidth,
xlabel={\# unlabeled data $N_u$},
xmajorgrids,
xtick style={color=black},
ylabel={$\LS$},
ymajorgrids,
y label style={at={(axis description cs:+0.14,0.93)},rotate=-90,anchor=south},
]
\addplot [thick, color0, mark=*, mark size=2, mark options={solid}]
table {%
0 -21.8085345219875
32 -14.2406162143173
64 -12.2864430350192
128 -11.5625555892574
};
\addlegendentry{LS}


\nextgroupplot[
height=\figHeight,
legend cell align={left},
legend style={fill opacity=0.8, draw opacity=1, text opacity=1, at={(0.97,0.03)}, anchor=south east, draw=white!80!black},
tick align=outside,
tick pos=left,
title={\# labeled data $N_l = 64$},
width=\figWidth,
xlabel={\# unlabeled data $N_u$},
xmajorgrids,
xtick style={color=black},
ylabel={$\LS$},
ymajorgrids,
y label style={at={(axis description cs:+0.14,0.93)},rotate=-90,anchor=south},
]
\addplot [thick, color0, mark=*, mark size=2, mark options={solid}]
table {%
0 0.576718978640357
32 1.04825359890642
64 1.24973018340608
128 1.5146939699889
};
\addlegendentry{LS}


\end{groupplot}
\end{tikzpicture}
    \caption{The predictive performance of the generative model as a function of the number of unlabeled data $N_u$ for $N_l = \left\lbrace 32, 64 \right\rbrace$. \resamplingstatement }
    \label{fig:effect_unlabeled_data}
\end{figure}
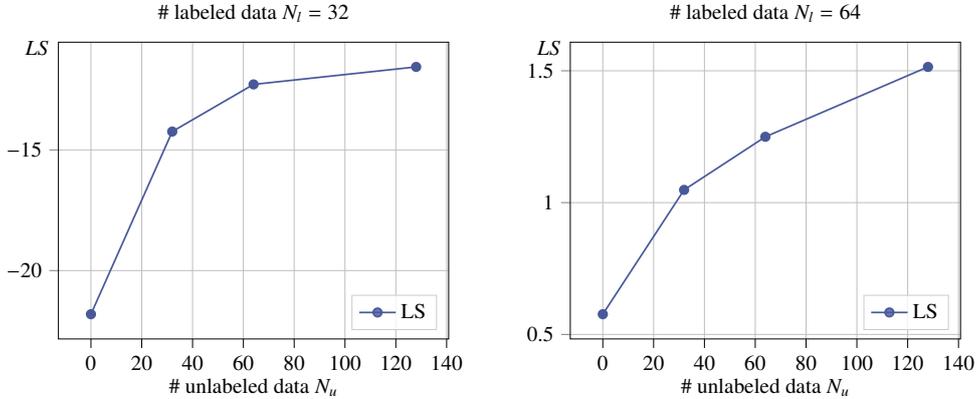


\subsection{Effect of the lower-dimensional encoding  and the CGM}
\label{sec:effectxx}

In the following we provide a brief exposition of the effect of the dimension of the latent encoding $\bz$  and  the state variables $\bxx$ (and $\byy$) on the predictive accuracy. In Figure \ref{Fig:study_latentdim_CGM} we alter the dimension of the $\text{dim} \left( \vect{z} \right)$ and clearly observe the existence of the information bottleneck; i.e. there exists threshold for $dim(\bz)$ up to which an improvement of the generative model is observed (for a given number of labeled data $N_l = 256$ and $N_u = 256$). After this threshold the predictive capability of the model deteriorates, since the ability to retain more information in the latent encodings is now superseded by the inability of the model to generalize well in the low-data-regime about the mappings linking the latent space to effective properties $\mat{X}$ and random field discretizations $\vect{x}$.

For the resolution of the CGM (or correspondingly the dimension of $\vect{X}$) one would assume to see an improvement in performance, as long as the dimension of the latent space as well as the number of datapoints afford the ability to exploit the increasing expressibility of the CGM. Here for a $\left( 32 \times 32 \right)$ FGM, $ \text{dim} \left( \vect{z} \right) = 32$, $N_l = 512, N_u = 512$ we illustrate in Figure \ref{Fig:study_CGM} the improvement of the predictive performance as the discretization of the CGM is increased from a $\left( 1 \times 1 \right)$ to a $\left( 4 \times 4 \right)$ . Supplementing these results we showcase in Figure \ref{fig:coarse_graining_illustration} how effective properties of a CGM change as the dimension of $\vect{X}$ increases.

\setlength{\figHeight}{5.0cm}
\setlength{\figWidth}{6.0cm}

\begin{figure}[h!]
 \captionsetup[subfigure]{width=0.9\linewidth}
\centering
\subcaptionbox{Predictive performance as a function of $\text{dim} \left( \vect{X} \right)$ ($N_l = 512, N_u = 512, N_\vo = 0, Q=32$). \resamplingstatement \label{Fig:study_CGM}}
{
\begin{tikzpicture}

\pgfplotsset{
  grid style = {
    dash pattern = on 0.05mm off 1mm,
    line cap = round,
    black,
    line width = 0.5pt
  }
}

\begin{axis}[
xmode = log,
log basis x={2},
grid=major,
height=\figHeight,
tick align=outside,
tick pos=left,
width=\figWidth,
x grid style={white!69.0196078431373!black},
xlabel={$\text{dim} \left( \bm{X} \right)$},
xmin=2, xmax=36,                                           
xtick style={color=black},
xtick = {},
xticklabels = {},
y grid style={white!69.0196078431373!black},
ylabel={\textcolor{color0}{$\LS$}},
y label style={at={(axis description cs:+0.14,0.93)},rotate=-90,anchor=south},
ytick style={color=color0},
y tick label style={color=color0},
y axis line style={color=color0},
]
\addplot [thick, color0, mark=*, mark size=2, mark options={solid}]
table {%
2 1.32466914653778                  
8 2.10146975517273                    
32 2.62939686775208
}; \label{LogscorePlot}

\end{axis}

\begin{axis}[
xmode = log,
log basis x={2},
axis y line=right,
height=\figHeight,
tick align=outside,
width=\figWidth,
x grid style={white!69.0196078431373!black},
xmin=2, xmax=36,                                           
xtick pos=left,
xtick style={color=black},
xtick = {2,8,32},                                          
xticklabels = {2, 8, 32},
y grid style={white!69.0196078431373!black},
ylabel={\textcolor{color1}{$R^2$}},
y label style={at={(axis description cs:+0.9,0.93)},rotate=-90,anchor=south},
ytick pos=right,
ytick style={color=black},
ytick style={color=color1},
y tick label style={color=color1},
y axis line style={color=color1},
legend pos=south east,
]
\addplot [dashed,thick, color1, mark=square*, mark size=2, mark options={solid}]
table {%
2 0.771464323997498
8 0.961793732643127
32 0.985347330570221
}; 

\end{axis}

\end{tikzpicture} 
}
\subcaptionbox{Predictive performance as a function of the dimension of the latent space dimension $Q = \text{dim} \left( \vect{z} \right)$; bottleneck occurs after $\text{dim} \left( \vect{z} \right) = 32$ ($\text{CGM} = \left( 4 \times 4 \right)$, $N_\labeled = 256$, $N_\unlabeled = 256$). \resamplingstatement  \label{Fig:study_latentdimt}}
{
\begin{tikzpicture}

\pgfplotsset{
  grid style = {
    dash pattern = on 0.05mm off 1mm,
    line cap = round,
    black,
    line width = 0.5pt
  }
}

\begin{axis}[
xmode = log,
log basis x={2},
grid=major,
height=\figHeight,
tick align=outside,
tick pos=left,
width=\figWidth,
x grid style={white!69.0196078431373!black},
xlabel={$\text{dim} \left( \bm{z} \right)$},
xmin=-1.1, xmax=67.1,
xtick style={color=black},
xtick = {},
xticklabels = {},
y grid style={white!69.0196078431373!black},
ylabel={\textcolor{color0}{$\LS$}},
y label style={at={(axis description cs:+0.14,0.93)},rotate=-90,anchor=south},
ymin=-9.98298953326792, ymax=3.16803528117388,
ytick style={color=color0},
y tick label style={color=color0},
y axis line style={color=color0},
]
\addplot [thick, color0, mark=*, mark size=2, mark options={solid}]
table {%
2 2.00690379142761
4 2.11108530044556
8 2.28168393135071
16 2.51330949783325
32 2.67397585868835
64 -7.07007142871618
}; \label{LogscorePlot}

\end{axis}

\begin{axis}[
xmode = log,
log basis x={2},
axis y line=right,
height=\figHeight,
tick align=outside,
width=\figWidth,
x grid style={white!69.0196078431373!black},
xmin=-1.1, xmax=67.1,
xtick pos=left,
xtick style={color=black},
xtick = {2,4,8,16,32 ,64},
xticklabels = {2,4,8, 16, 32, 64},
y grid style={white!69.0196078431373!black},
ylabel={\textcolor{color1}{$R^2$}},
y label style={at={(axis description cs:+0.9,0.93)},rotate=-90,anchor=south},
ymin=0.948538049161434, ymax=0.990845804512501,
ytick pos=right,
ytick style={color=black},
ytick={0.94, 0.95, 0.96, 0.97, 0.98},
yticklabels={0.94, 0.95, 0.96, 0.97, 0.98},
ytick style={color=color1},
y tick label style={color=color1},
y axis line style={color=color1},
legend pos=south east,
]
\addplot [dashed,thick, color1, mark=square*, mark size=2, mark options={solid}]
table {%
2 0.951175317764282
4 0.960582091808319
8 0.971398632526398
16 0.982615456581116
32 0.989779362678528
64 0.974454171657562
}; 

\end{axis}

\end{tikzpicture} 
}
\caption{Effect of the dimension of the latent encoding $\vect{z}$ and $\vect{X}$ on the predictive performance}\label{Fig:study_latentdim_CGM}
\end{figure}
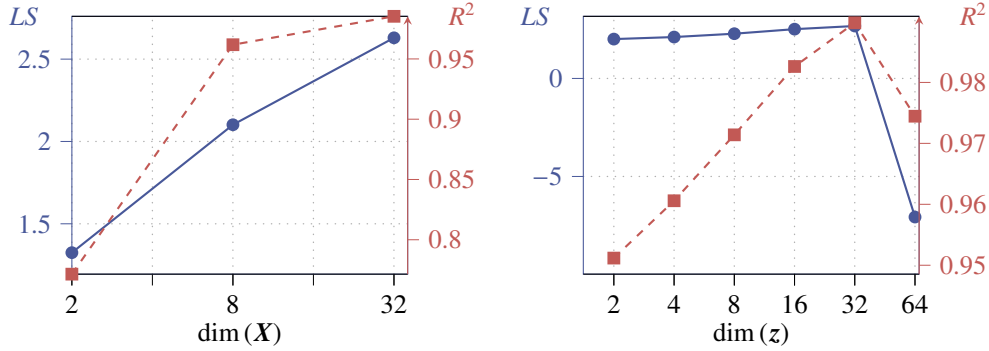


\subsection{Effect of different BCs}
\label{sec:effectbc}

In the following we evaluate the predictive performance of the model in an \textit{extrapolative} setting, i.e. when the model is asked to provide predictions for boundary conditions not observed during training. To this end we consider the set of boundary conditions listed in Table \ref{tab:cross_pred_performance_boundary_configs}, where the coefficients $a_i$ refer to the definition of a parametric Dirichlet B.C. as given in Equation \eqref{Eq:randomized_dirichlet_bc} (for any $a_i$ we specify either a fixed value, or a distribution of it to be randomly sampled from).

\begin{table}[h!]
\begin{subtable}{.5\linewidth}\centering
{
\begin{small}
\begin{center}
    \textbf{Boundary Conditions}
\end{center}
\begin{tabular}{c rrrrr} 
\hline 
& A & B & C & D \\ [0.5ex]
\hline \hline 
$a_0$ & $0$ & $1$ & $\mathcal{U} \left( -0.5, 0.5 \right)$ & $0$ \\ 
$a_1$ & $0$ & $1$ & $0$                                    &  $\text{Beta} \left( 2, 5 \right)$ \\ 
$a_2$ & $1$ & $0$ & $0$                                    &  $-\text{Beta} \left( 2, 5 \right)$ \\ 
$a_3$ & $1$ & $0$ & $\mathcal{U} \left( -0.5, 0.5 \right)$  & $0$ \\[1ex] 
\hline 
\end{tabular}
\vspace{2mm}
\end{small}
}
\caption{} \label{tab:cross_pred_performance_boundary_configs}
\end{subtable}%
\begin{subtable}{.5\linewidth}\centering
{
\begin{small}
\begin{center}
    \textbf{Logscore} $\LS$
\end{center}
\begin{tabular}{r||c|c|c|c}
\begin{tabular}{c} prediction on \\ \hline trained on \end{tabular}
& A & B & C & D \\\hline\hline
A & 1.30 & 1.30 & 2.61 & 2.34  \\\hline
B & 1.40 & 1.40 & 2.64 & 2.39 \\ \hline
C & 1.26 & 1.24 & 2.75 & 2.30 \\ \hline
D & 1.17 & 1.13 & 2.44 & 2.42 
\end{tabular}
\vspace{2mm}
\end{small}
}
\caption{} \label{tab:cross_pred_performance_logscore}
\end{subtable}
\caption{(a) Different BCs considered, and (b) Predictive performance $LS$ score obtained when training a model under the BCs indicated by the row and tested on the BCs indicated by the column.}
\label{tab:cross_pred_performance}
\end{table}

In Table \ref{tab:cross_pred_performance_logscore} we report the $LS$ score obtained on a validation dataset ($N_v = 256$) In all cases the model was trained on $N_l = 512$ labeled and $N_u = 2048$ unlabeled data (with  $N_\vo = 0$) using an amortized encoder. The diagonal terms correspond to predictive scores on the same BCs as the ones used for training (interpolative), whereas the off-diagonal ones to scores obtained on different BCs than the ones used for training (extrapolative). 
The results indicate that the predictive performance does not significantly depend upon the type of boundary condition the model has been trained on, i.e. the predictive performance in Table \ref{tab:cross_pred_performance_logscore} only varies marginally across a column (BC used for training), and the  variation is  mostly determined (see row-wise), on which kind of boundary conditions we wish to make predictions.


\subsection{Application: Uncertainty Propagation}
\label{subsection:uq}

\begin{figure}[h!]
\begin{center}
\input{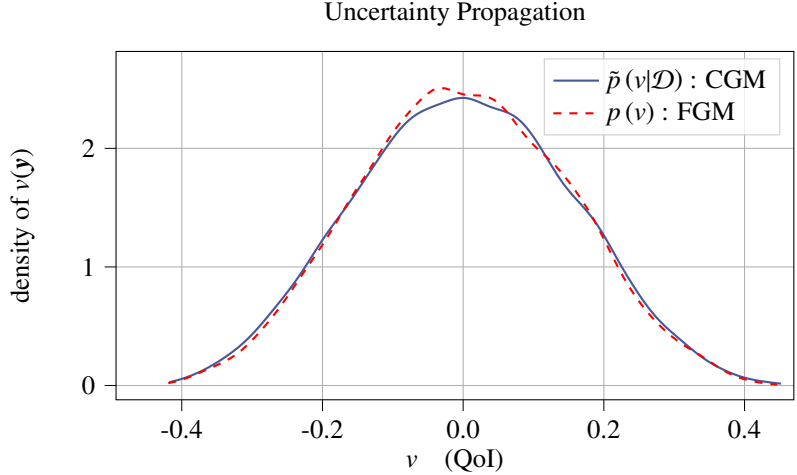}
\end{center}
\caption{The predictive posterior density $p\left( v  | \mathcal{D}\right)$ over the  QoI  $v(\by)$ as compared with the Monte Carlo reference $p(v)$ obtained with $N_{MC} = 8192$ FGM solves.
The model has been trained using   $N_\labeled = 32$ (compare this with $N_{MC}$), $N_\unlabeled = 8192$ and $N_{\vo} = 256$ \textit{Hybrid} virtual observables (see section \ref{subsec:study_virtual_obs}) and an amortized encoder was used for predictions. }
\label{Fig:UQ_DensityComparison}
\end{figure}

As mentioned earlier, many-query applications represent one of the main incentives for learning such probabilistic surrogates.
We consider here the case of uncertainty propagation where the goal is to compute statistics of Quantities of Interest (QoIs)  associated with the output $\by$ when the input $\bx$ is random with a density, say $p \left( \vect{x} \right)$. In the sequel, we compare the reference solution for the density of such a  scalar QoI $v(\by)$ obtained by direct Monte Carlo (i.e. by generating $N_{MC}= 8192 $) samples of $\bx$ and solving $N_{MC}$ times the FGM) with the marginal distribution $\tilde{p} \left( v | \mathcal{D} \right)$ over the QoI  obtained from the posterior predictive as follows:

\begin{align}
\tilde{p} \left( v | \mathcal{D} \right) = \int \int  \delta \left( v - v\left( \vect{y} \right) \right) p \left( \vect{y} \middle| \vect{x} , \mathcal{D} \right) p \left( \vect{x} \right) \diff \vect{x} \diff \vect{y}
\label{Eq:MarginalUQ_distribution}
\end{align}
We chose as $v(\by)$ the value of the solution of the PDE at the middle of our computational domain, i.e. at $s=(0.5,0.5)$.
The generative model was trained with $N_\unlabeled = 8192, N_\labeled = 32$ and $N_\vo = 256$ 
 and the results obtained are illustrated in Figure \ref{Fig:UQ_DensityComparison}. The approximation $\tilde{p} \left( v  | \mathcal{D} \right)$ obtained from the probabilistic surrogate matches closely with the Monte Carlo reference. If we had adopted a fully Bayesian approach, i.e. if $p \left( \vect{\theta} \middle| \mathcal{D} \right)$ was captured beyond a point estimate, additional uncertainty bounds on the probability density function $\tilde{p} \left( v | \mathcal{D} \right)$ could be derived \citep{schoberl2019predictive}. Note that the approximate marginal distribution $\tilde{p} \left( v  | \mathcal{D} \right)$ has been obtained by leveraging the amortized encoder $p_{\Phi} \left( \vect{z} \middle| \vect{x} \right)$, such that each prediction merely requires to pass $\vect{x}$ through a neural network and to solve the CGM.

\vspace{0.5cm}

\setlength{\figHeight}{6.2cm}
\setlength{\figWidth}{10.5cm}


\section{Conclusions}
\label{sec:conclusions}

We have proposed a generative probabilistic model for constructing surrogates for PDEs characterized by high-dimensional parametric inputs $\bx$ and high-dimensional outputs $\by$. Its most important and novel characteristics are:
\bi
\item it learns the joint density $p(\bx, \by)$ in contrast to the conditional $p(\by| \bx)$ that most {\em discriminative} models in the literature target. As a result, it can make use of {\em unlabelled} data (i.e. only inputs $\bx$) in a semi-supervised fashion.
\item it employs a supervised dimensionality reduction scheme that identifies a set of lower-dimensional features $\bz$  that are most predictive of the output $\by$. The training of this component is integrated in the overall model and is performed simultaneously with the rest.
\item it employs a coarse-grained model at its core which serves as an information bottleneck between the high-dimensional inputs $\bx$ and outputs $\by$. We have demonstrated how such models can be flexibly constructed by coarsening the FGM and have shown  that this can lead to superior predictive performance in the {\em Small labeled Data} regime as well as under extrapolative conditions (i.e. boundary conditions {\em not} used during training). 
\item it makes use of domain knowledge in the form of constraints/equalities or functionals that govern the original physical problem. These are incorporated in the likelihood in a fully Bayesian fashion as {\em virtual observables} and can lead to significant performance gains while reducing further the need for expensive, labeled data.
\item it yields a predictive posterior density that can be used not only for point estimates, but  for quantifying the predictive uncertainty as well. The latter is most often  neglected in similar efforts but it is an unavoidable consequence of any coarse-graining or dimensionality-reduction or reduced-order-modeling scheme that is trained on finite amounts of data.

\ei

The proposed  modeling framework provides  a fertile ground for several extensions. Apart from the obvious refinement, both in terms of breadth and depth, of the neural nets employed these improvements would involve:
\bi
\item the automatic discovery of the dimension of the latent variables $\bz$ as well as of the CGM. In the latter case, this could involve  the dimension of the state variables $\bxx,\byy$ as well as the model-form itself i.e. the relation between $\bxx$ and $\byy$. As previously mentioned, the ELBO $\mathcal{F}$  could serve as the driver for such investigations since it quantifies the plausibility of the data under  a given model  by balancing the quality of the fit with the model's complexity \citep{rasmussen_c.e._occams_2001,grigo_physics-aware_2019}.
\item active learning in terms of  unlabeled data  and virtual observables. As it has been demonstrated, such  data provide valuable information in improving the model. It is not necessary though that all inputs $\bx$ or pairs of inputs and virtual observables $(\bx,\hat{\bs{o}})$ provide the same information. A critical component in improving the overall training efficiency would be to employ active learning schemes \citep{kandasamy_query_2017} in order to adaptively select the inputs and/or virtual observables (e.g. weight functions) at each step that are most informative. We note that such a scheme and in the context of a {\em deterministic} PDE-surrogate has been proposed in    \citep{khodayi-mehr_varnet:_2019}. Extensions in the probabilistic setting advocated  could also make use of the ELBO in selecting from a vocabulary of options, the one that would lead to the largest increase in $\mathcal{F}$.

\ei


\clearpage
 \appendix

\section{Encoding Conservation laws as equality constraints} 
\label{App:FluxConstraint}
\begin{figure}[!ht]
\begin{center}
\def\svgwidth{\columnwidth}
\begin{Large}
\resizebox{0.55\textwidth}{!}{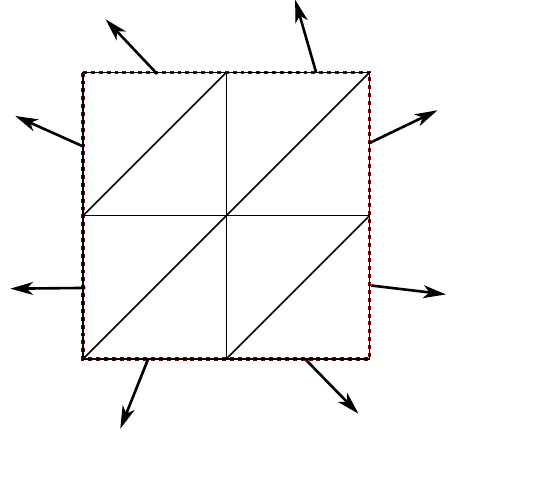}
\end{Large}
\caption{If the source term $f_i$ associated with subdomain $\Omega_i$ is zero, then the integrated flux across the boundary should net to zero. The discrepancy of this flux $o_i := \Delta \Psi_{\Omega_i}$ corresponds to a virtual observable (equality constraint) introduced as artificial node in our probabilistic graphical model. }
\label{Fig:FluxIllustration}
\end{center}
\end{figure}

 A wide range of PDEs imply physical conservation laws, i.e. the governing equation state that some quantity $\Psi$ is conserved and unchanging.  Since this holds for any arbitrary subdomain $\Omega_i \subset \Omega$ and time interval we may express this in integral form \citep{lee2019deep} as

\begin{align}
\Delta \Psi_{\Omega_i} \left( t \right) = \frac{\diff}{\diff t} \int_{\Omega_i} \int \Psi \left( \vect{s}, t \right) \diff \Omega_i + \int_{\partial \Omega_i} \flux_i \left( \vect{s}, t \right) \diff \left( \partial \Omega_i  \right) - \int_{\Omega_i} f_i \left( \vect{s}, t \right) \diff \Omega_i
\label{Eq:ConservationGeneralForm}
\end{align}

where $\vect{s}$ , $\flux_i$ and $f_i$ denote the spatial coordinates, (boundary) flux and source term of subdomain $\Omega_i$, respectively. We may introduce this physical conservation constraint into our model by introducing $o_i = \Delta \Psi_{\Omega_i}$ as a virtual observable. A virtual observable may then for instance correspond to violation of energy conservation resulting from the CGM predictions, entering into the probabilistic model by virtue of a zero-mean virtual Gaussian likelihood (e.g. $\smash{o_i := \Delta \Psi_{\Omega_i} \sim \mathcal{N} \big( 0, \tau_i^{-1} \big)}$. For our steady-state elliptic problem with no time-dependence  \refeqp{Eq:ConservationGeneralForm} simplifies

\begin{align}
\Delta \Psi_{\Omega_i} = \int_{\partial \Omega_i}  \flux_i \left( \vect{s} \right) \diff  \Gamma - \int_{\Omega_i}  f_i \left( \vect{s} \right)  \cdot \diff \Omega_i 
\label{Eq:FluxSubdomain}
\end{align}

which states that the net-flow across the boundary $\partial \Omega_i$ must be equal to production specified by the source term (see also \refeqp{Eq:Poisson0}). With $u = \sum_{j=1}^{d_y} \varphi_j^u \left( \vect{s} \right) y_j$ given by a Finite Element discretization of local (linear) shape functions defined on some triangulation $\mathcal{T}$ of the computational domain,  \refeqp{Eq:FluxSubdomain} results in a linear constraint, since the flux is element-wise constant (see Figure \ref{Fig:FluxIllustration}), enabling us to compute

\begin{align}
\int\limits_{\partial \Omega_i} \flux \left( \vect{s} \right) \diff \Gamma = \sum\limits_{j=1}^{N_e} \vect{n}_{e_j}^T \flux_{e_j} 
\label{Eq:js9j3}
\end{align}

where the element-wise constant flux $\flux_{e_i} = \mat{B}^{(i)} \vect{y}$ is linear in $\vect{y}$ with $\mat{B}^{(i)} \in \mathbb{R}^{2 \times d_y}$, and we sum over all finite elements comprising the subdomain $\Omega_i$ (assuming a compliant mesh). As such for the choice of $M$ subdomains $\Omega_i, i=1, ..., M$ we may define as virtual observable a vector $\vect{o} \left( \vect{y}; \vect{x} \right)$ (where the $i$-th entry corresponds to $\Delta \Psi_{\Omega_i}$) which can be expressed as

\begin{align}
\vect{o} \left( \vect{y}; \vect{x} \right) = \mat{\Gamma} \left( \vect{x} \right) \vect{y} - \vect{\alpha} \left( \vect{x} \right) 
\end{align}

with the entries of $\mat{\Gamma} \left( \vect{x} \right)$ deriving from \eqref{Eq:js9j3} and $\vect{q}_{e_i} = \mat{B}^{(i)} \vect{y}$, while $\alpha_i = \int_{\Omega_i} f_i \left( \vect{s} \right) \cdot \diff \Omega_i$.

\section{Low-Rank Mean-Field updates for virtual observables}
\label{sec:virtual_obs_closed_form}

While in principle the entire model can be trained using stochastic variational inference \footnote{The required Jacobian of the virtual observables $\vect{o} \left( \vect{y}, \vect{x} \right)$ in order to propagate gradients simply reduces to the well-known Gateaux derivative, and is easily (as well as cheaply and parallelizeable) obtained in most Finite Element frameworks (see e.g. \textit{Unified Form Language} \citep{alnaes2014unified})} as outlined in Algorithm \ref{Alg:GeneralAlg}, for linear equality constraints we are able to perform closed-form mean-field updates for $q \left( \mathcal{Y}_\mathcal{O} \right)$, providing both additional insight as well as computationally efficient updates. For any ensemble of linear physical constraints enforced with a certain precision $\mat{\Lambda}$ we may write

\begin{align}
\vect{o}  \left( \vect{y} , \vect{x} \right) := \mat{\Gamma} \left( \vect{x} \right) \vect{y} - \vect{\alpha} \left( \vect{x} \right) \sim \mathcal{N} \left( \vect{0}, \mat{\Lambda}^{-1} \right) \qquad \qquad \mat{\Gamma} \left( \vect{x} \right)= \left[ \vect{\gamma}_1 \left( \vect{x} \right)^T, ...,  \vect{\gamma}_M \left( \vect{x} \right)^T \right] \in \mathbb{R}^{M \times d_y} 
\label{Eq:AppendixGenLinearObsRepeat}
\end{align}

where the entries of $\mat{\Gamma} \left( \vect{x} \right)$ and $\vect{\alpha} \left( \vect{x} \right)$ derive from the particular choice of constraint and the underlying physics at a query point $\vect{x}$ (see section \ref{sec:Example_VO}). The precision matrix $\mat{\Lambda} = \text{diag} \left( \lambda_1, ..., \lambda_M \right)$ is chosen diagonal, such that the set of parameters $\tau$ governing the enforcement of our constraints follows as $ \tau = \big\lbrace \lambda_i \big\rbrace_{i=1}^M$. Given the assumed structure of the variational approximation $q_{\vect{\xi}} \left( \vect{\theta}, \mathcal{R} \right)$ (see \refeqp{Eq:VariationalFactorization}), note that the optimal $q^* \left( \mathcal{Y}_{\mathcal{O}} \right)$ follows by integrating out all other factors of $q_{\vect{\xi}}$ \citep{bishop_pattern_2006}

\begin{align}
\log q^* \left( \mathcal{Y}_{\mathcal{O}} \right) &= 
\mathbb{E}_{\tilde{q}_{\vect{\xi}}} \left[ \log \left( p \left( \hat{\mathcal{O}} \middle| \mathcal{Y}_{\vo}, \mathcal{X}_{\vo}, \mat{\Lambda} \right) p \left( \mathcal{Y}_{\vo} \middle| \mathfrak{X}_{\vo}, \vect{\theta} \right) p \left( \mathfrak{X}_{\vo} \middle| \mathcal{Z}_{\vo} , \vect{\theta} \right) 
p \left( \mathcal{X}_{\vo} \middle| \mathcal{Z}_{\vo}, \vect{\theta}  \right)
p \left( \mathcal{Z}_{\vo} \right)
p \left( \vect{\theta} \right) 
\right)  
\right]  \nonumber
\\
 &= \mathbb{E}_{\tilde{q}_{\vect{\xi}}} \left[ 
 - \sum\limits_{i_\vo=1}^{N_{\mathcal{O}}} \left[ 
  \frac{1}{2} \left( \vect{y}^{(i_\vo)} - \vect{h} \left( \mat{X}^{(i_\vo)} \right) \right)^T \mat{S}_{\vect{y}}^{-1}  \left( \vect{y}^{(i_\vo)} - \vect{h} \left( \mat{X}^{(i_\vo)} \right) \right) 
  \right]
 \right]  \nonumber \\
 \phantom{=E} &+ \mathbb{E}_{\tilde{q}_{\vect{\xi}}} \left[ - \sum\limits_{i_\vo=1}^{N_{\mathcal{O}}}
 \left[ \frac{1}{2}
\left(  \mat{\Gamma} \left( \vect{x}^{(i_\vo)} \right)  \vect{y} - \vect{\alpha} \left( \vect{x}^{(i_\vo)} \right) \right)^T \mat{\Lambda} \left(  \mat{\Gamma} \left( \vect{x}^{(i_\vo)} \right)  \vect{y} - \vect{\alpha} \left( \vect{x}^{(i_\vo)} \right) \right)
 \right] \right] + \text{const.} \label{Eq:jasdjgsd}
\end{align}

where $\hat{\mathcal{O}} = \big\lbrace \vect{\hat{o}} \big\rbrace_{i_\vo=1}^{N_\vo} $ comprises all virtual observations and  $\tilde{q}_{\vect{\xi}}$ denotes all other factors of the structured mean-field approximation aside from $q \left( \mathcal{Y}_{\vo} \right)$, i.e. $q_{\vect{\xi}} = q \left( \mathcal{Y}_\vo \right) \tilde{q}_{\vect{\xi}} $. Inspecting \refeqp{Eq:jasdjgsd} we find that it is linear-quadratic in $\vect{y}$, which implies a Gaussian $q \big( \vect{y}^{(i_\vo)} \big) = \mathcal{N} \big( \vect{\mu}^{(i_\vo)}, \mat{\Sigma}^{(i_\vo)} \big)$ at every query point with  mean and covariance implicitly defined by (for $i_\vo=1, ..., N_{\mathcal{O}}$)

\begin{align}
\left. \mat{\Sigma}^{(i_\vo)} \right.^{-1} \vect{\mu}^{(i_\vo)} &= \mat{\Gamma} \left( \vect{x}^{(i_\vo)} \right)^T \mat{\Lambda} \left( \vect{x}^{(i_\vo)} \right) \vect{\alpha} \left( \vect{x}^{(i_\vo)} \right) + \left\langle \mat{S}_{\vect{y}}^{-1} \right\rangle \left\langle \vect{h} \left( \vect{Y} \left( \mat{X}^{(i_\vo)} \right); \vect{\theta} \right) \right\rangle  \nonumber \\
 \left. \mat{\Sigma}^{(i_\vo)} \right. ^{-1} &= \mat{\Gamma} \left( \vect{x}^{(i_\vo)} \right)^T  \mat{\Lambda}  \mat{\Gamma} \left( \vect{x}^{(i_\vo)} \right) + \left\langle \mat{S}_{\vect{y}}^{-1} \right\rangle
\label{Eq:ClosedFormGaussianVO}
\end{align}

where $\left\langle \cdot \right\rangle$ denotes an expectation with respect to all remaining factors of the variational approximation $\tilde{q}_{\vect{\xi}}$. Given our model choices (Eqs.\eqref{Eq:model_0} - \eqref{Eq:model_C}), the expectation of the precision matrix  $\smash{\left\langle \mat{S}_{\vect{y}}^{-1} \right\rangle}$ is constrained to be diagonal  while the matrix $\smash{\mat{\Gamma} \big( \vect{x}^{(i)} \big)^T  \mat{\Lambda}  \mat{\Gamma} \big( \vect{x}^{(i)} \big)}$ with $\mat{\Gamma} \in \mathbb{R}^{M \times d_y}$ exhibits low-rank structure. This low-rank structure reflects the fact that we only have introduced \textit{partial} or \textit{incomplete} information, and as such the constraints are only informative for a certain (low-dimensional) subspace. It simultaneously allows us to cheaply incorporate this physical knowledge into our model, since we may exploit the low-rank structure and use the Woodbury matrix identity to obtain mean vector and covariance matrix of the Gaussians $q \big( \vect{y}^{(i_\vo)} \big) = \mathcal{N} \big( \vect{\mu}^{(i_\vo)}, \mat{\Sigma}^{(i_\vo)} \big)$ at a cost $\mathcal{O} \big( M^3 \big)$, i.e. numerical expense of updating $\smash{q \big( \vect{y}^{(i)} \big)}$ depends on the number of enforced constraints rather than the dimension of $\vect{y}$.  Making use of the Woodbury matrix identity one finds

\begin{align}
\mat{\Sigma}^{(i_\vo)} &=  \left\langle \mat{S}_{\vect{y}} \right\rangle - \left\langle \mat{S}_{\vect{y}} \right\rangle \mat{\Gamma} \left( \vect{x}^{(i_\vo)} \right)^T  \left. \mat{\Xi}^{(i_\vo)} \right.^{-1}  \mat{\Gamma} \left( \vect{x}^{(i_\vo)} \right) \left\langle \mat{S}_{\vect{y}} \right\rangle 
\label{Eq:qy_closed_form_update_woodbury}
\end{align}

where we have introduced the $M \times M$ matrix $\mat{\Xi}^{(i_\vo)} = \mat{\Gamma} \big( \vect{x}^{(i_\vo)} \big) \left\langle \mat{S}_{\vect{y}} \right\rangle \mat{\Gamma} \big( \vect{x}^{(i_\vo)} \big)^T + \mat{\Lambda}^{-1}$. In the limit case of components of the diagonal precision matrix $\mat{\Lambda}$ being infinite (i.e. absolute enforcement of the constraint), the result is an am improper Gaussian with rank-deficient covariance, i.e. the epistemic uncertainty of the epistemic uncertainty of the model collapses to a subspace which is completely in compliance with the enforced constraints; the update of $q \left( \mathcal{Y}_\vo \right)$ then becomes similar to the updates of Bayesian Conjugate Gradient (BCG) \citep{cockayne2018bayesian}, which poses the solution of a linear equation system as a problem of probabilistic inference conditionally on the observance of a set of search directions.

\section{Adaptively inferring finite precisions}
\label{sec:FinitePrecisionUpdates}

For some physical constraints as e.g. the flux constraint (\ref{App:FluxConstraint}) it is neither plausible to assume infinite precision, nor do we a-priori know a suitable finite precision value with which to enforce the constraint. In such cases we may chose to treat the precision parameters $\tau = \left\lbrace \lambda_m \right\rbrace_{m=1}^{M} $ probabilistically as well. We propose to introduce a Gamma prior $\lambda_i \sim \Gamma \left( \alpha_0^{(i)}, \beta_0^{(i)} \right)$ for each of the unknown precision values $\lambda^{(i)}$, or alternatively assume identical precision for all virtual observables (or subgroups thereof). For notational simplicity we discuss the latter case where all virtual observables are governed by a singular precision parameter $\lambda$

\begin{align}
\lambda \sim \frac{\beta_0^{\alpha_0}}{\Gamma \left( \alpha_0 \right)} \lambda^{\alpha_0 -1} \exp \left( - \beta_0 \lambda \right)
\end{align}

The variational approximation is extended to include $q \left( \lambda \right)$, and identically to the closed-form updates of $q \left( \mathcal{Y}_{\mathcal{O}} \right)$ in  \ref{sec:virtual_obs_closed_form} the optimal variational approximation $q^* \left( \lambda \right)$ is to be found a Gamma distribution  $\Gamma \left( \alpha, \beta \right)$, with parameters $\alpha$ and $\beta$ given by

\begin{align}
\alpha = \left( \sum\limits_{i_\vo=1}^{N_{\vo}} \frac{1}{2} M \right) + \alpha_0 \qquad \qquad \beta = \frac{1}{2} \sum\limits_{i_\vo=1}^{N_{\mathcal{O}}}  \mathbb{E}_{q \left( \vect{y}^{(i_\vo)} \right)} \left[ \left| \left| \vect{o} \left( \vect{y}^{(i_\vo)} ; \vect{x}^{(i_\vo)} \right) \right| \right|_2^2 \right] + \beta_0
\label{Eq:PrecisionGammaMeanField}
\end{align}

where $M$ the number of constraints at each query point governed by $\lambda$. For a linear constraint \eqref{Eq:AppendixGenLinearObsRepeat} and a Gaussian $\smash{q  \big( \vect{y}^{(i_\vo)} \big) = \mathcal{N} \big( \vect{\mu}^{(i_\vo)}, \mat{\Sigma}^{(i_\vo)} \big)}$ as given by Equation \eqref{Eq:ClosedFormGaussianVO} the expectation involved in finding $\beta$ becomes tractable; otherwise they can be cheaply estimated using Monte Carlo. For the Gamma prior we chose $\alpha_0 = \beta_0 = 10^{-6}$. 
 
\section{Stochastic Second Order Optimization for the energy-based virtual observables}
\label{sec:PotFctVirtObs}

The introduction of the energy as a virtual observable at $N_{\vo}$ query point differs from the other constraints we considered, since in contrast to $M << d_y$ equality constraints it \textit{fully} summarizes all the information about the governing equations. Specifically,  for a Finite Element discretization of the linear elliptic PDE given by $\mat{K} \left( \vect{x} \right) \vect{y} = \vect{f} \left( \vect{x} \right)$, the energy can be expressed in discretized form as

\begin{align}
    V \left( \vect{y}^{(i_\vo)}, \vect{x}^{(i_\vo)} \right) = \frac{1}{2} \left. \vect{y}^{(i_\vo)} \right. ^T \mat{K} \left( \vect{x}^{(i_\vo)} \right) \vect{y}^{(i_\vo)} - \vect{f} \left( \vect{x}^{(i_\vo)} \right) ^T \vect{y}^{(i_\vo)}
\end{align}

and we find that the minimization of the quadratic potential $V \left( \vect{y}^{(i_\vo)}, \vect{x}^{i_\vo} \right)$ is the dual problem to solving the linear equation system associated with the solution of the discretized PDE itself. The introduction of the energy similarly implies that the ELBO becomes a quadratic potential in $\vect{\mu}^{(i_\vo)}$; i.e. plausibility of the model as scored by the ELBO now depends on the energy state obtained for predictions at all $N_\vo$ query points. Following the same mean-field approach as in \ref{sec:virtual_obs_closed_form}, the optimal $q \big( \vect{y}^{(i_\vo)} \big) = \mathcal{N} \big( \vect{\mu}^{(i_\vo)}, \mat{\Sigma}^{(i_\vo)} \big)$ is found to be a Gaussian with mean and covariance defined by (for $i= 1, ..., N_{\vo}$)

\begin{align}
\left. \mat{\Sigma}^{(i_\vo)} \right. ^{-1} \vect{\mu}^{(i_\vo)} =\tau  \vect{f}^{(i_\vo)} + 
 \left\langle \mat{S}_{\vect{y}}^{-1} \right\rangle \left\langle \vect{h} \left( \vect{Y} \left( \mat{X}^{(i_\vo)} \right); \vect{\theta} \right) \right\rangle
   \qquad \qquad \left. \mat{\Sigma}^{(i_\vo)} \right.^{-1} = \left\langle \mat{S}_{\vect{y}}^{-1} \right\rangle +  \tau \mat{K} \left( \vect{x}^{(i_\vo)} \right)
\label{Eq:PotentialClosedFormUpdate}
\end{align}

where $\tau$ is a precision or tempering parameter which governs the weight given to the virtual observables - for the limit case of $\tau$ approaching infinity, the belief about $\vect{y}^{i_\vo}$ will entirely depend on the energy state and becomes independent of the probabilistic surrogate. In contrast to the enforcement of $M << d_y$ equality constraint, the precision matrix $\left. \mat{\Sigma}^{(i_\vo)} \right.^{-1}$ is sparse but exhibits full-rank structure, precluding the possibility to perform low-rank updates. As such the maximization of the evidence lower bound as a quadratic potential w.r.t. $\vect{\mu}^{(i_\vo)}$ on first glance appears to be the dual problem to solving the linear PDE itself if no amortization is applied. Note however that 

\begin{itemize}
\item  the maximization of the ELBO defines a  simplified transfer problem since $\text{cond} \big( \tau \mat{K} ( \vect{x}^{(i_\vo)} ) + \big\langle \mat{S}_{\vect{y}}^{-1} \big\rangle \big) \leq \text{cond} \big( \mat{K} ( \vect{x}^{(i_\vo)} ) \big)$, i.e. the probabilistic surrogate implicitly acts as a preconditioner. When optimizing the evidence lower bound we merely use the energy to \textit{correct} the predictions of the surrogate and to pull them gradually in the right direction, instead of solving the PDE from scratch. This suggests an approach where one slowly tempers $\tau$ during training
\item knowledge is transferred and mediated by the probabilistic model, as opposed to solving $N_{\vo}$ entirely disjoint problems
\item we are not intrinsically interested in $q \big( \vect{y} \big)$ but only to the extend to which it is able to inform our probabilistic surrogate, (i.e. learn the parameters $\vect{\theta}$ of the generative model). As such, due to the inherent irreducible error introduced by the CGM, beyond a certain point there is no benefit in increasing $\tau$, which can be seen to correspond to the tolerance parameter of iterative solvers
\end{itemize}

Despite this it has to be noted that the incorporation of this inequality constraint is comparably much more expensive. Since we want to avoid solving the equation system implied by \refeqp{Eq:PotentialClosedFormUpdate} directly, we constrain the covariance matrix $\mat{\Sigma}^{(i_\vo)}$ of the variational approximation $q \big( \vect{y}^{(i_\vo)} \big) = \mathcal{N} \big( \vect{\mu}^{(i_\vo)}, \mat{\Sigma}^{(i_\vo)} \big)$ to be diagonal and chose to optimize $\mathcal{F}$ iteratively with respects to the parameters of $q \left( \vect{y}^{((i_\vo)} \right)$ using second order stochastic optimization. Here we use randomized Newton \citep{gower2015randomized, gower2019rsn}, which can be seen to iteratively updates parameters such that the iterates are as close as possible in the L2 norm,  while simultaneously forcing the error to be zero with respect to a randomly sampled subspace  (see \textit{sketching-viewpoint} of \citep{gower2015randomized}).


\newpage
\bibliographystyle{model6-num-names} 
\bibliography{references_mr,psk}

%
%
%

\end{document}